\documentclass{article} 

\usepackage{amsmath,amsfonts,bm}

\def\eqref#1{equation~\ref{#1}}

\def\1{\bm{1}}

\DeclareMathAlphabet{\mathsfit}{\encodingdefault}{\sfdefault}{m}{sl}
\SetMathAlphabet{\mathsfit}{bold}{\encodingdefault}{\sfdefault}{bx}{n}

\usepackage{iclr2027_conference,times}
\usepackage{hyperref}
\usepackage[capitalise]{cleveref}
\usepackage{xfrac}
\usepackage{url}
\usepackage[table]{xcolor}
\usepackage{graphicx}
\usepackage{float}
\usepackage{enumitem}
\usepackage{booktabs}
\usepackage{wrapfig}
 \usepackage{subcaption}
\usepackage{array}
\setlist[itemize]{leftmargin=*}

\iclrfinalcopy

\title{Stability Does Not Mirror Plasticity: \\A Module-Level View of Declarative and \\Procedural Learning in Transformers}

\author{Jinjun Peng$^{*}$, Weiliang Zhao$^{*}$, Baishakhi Ray, Zhou Yu, Junfeng Yang \\
Department of Computer Science\\
Columbia University\\
\texttt{\{jinjun,weiliang\}@cs.columbia.edu}
}

\begin{document}
\let\thefootnote\relax\footnotetext{* Equal Contribution}

\maketitle
\begin{abstract}
Plasticity and stability in continual learning are typically studied at the entire-model level and in a task-agnostic manner, yet prior work has reported divergent conclusions across settings.
Inspired by cognitive distinctions between \emph{declarative} and \emph{procedural} memory, we instead study how plasticity and stability vary across transformer modules and task types.
In a controlled synthetic environment with clean measurement, we show a double dissociation:
query-key (QK) and value-output (VO) projections are the most plastic on procedural tasks, whereas MLPs are the most plastic on declarative memorization.
Stability, however, does not simply mirror plasticity:
when learning a new procedure, updating QK retains over half of the stored declarative memory, while updating MLPs retains far less and updating VO loses roughly four-fifths.
Targeted module reset-and-recovery experiments explain this asymmetry: plasticity tracks which module is best suited to a task’s primary computation or storage, whereas stability depends on all ways the existing capability relies on that module, including the access to information stored in other modules.
Experiments on a realistic model and tasks broadly echo these asymmetric dynamics.
Overall, our work shows that plasticity and stability depend on different aspects of a module's functional role, offering a task-dependent, module-level perspective on continual learning.
\end{abstract}
\section{Introduction}

Existing continual learning study predominantly frames the plasticity (the ability to acquire new capabilities) and stability (the preservation of previously acquired knowledge) trade-off at the \textit{entire-model} level and in a \textit{task-agnostic} manner, seeking universal recipes including LoRA \citep{hu2022lora} and self-distillation \citep{yang2024selfdistillation,shenfeld2026selfdistillation}.
However, their empirical findings vary across adaptation settings.
Across tasks, while \citet{biderman2024lora} evaluate code and math observing LoRA ``learns less and forgets less,'' \citet{pletenev2025knowledge} find that injecting factual knowledge through LoRA still lead to noticeable degradation on external question-answering benchmarks.
Across architectural modules, \citet{schulman2025lora} show that attention-only LoRA underperforms MLP-only LoRA, whereas \citet{zhu2025teach} demonstrate that tuning only self-attention projections allows multimodal models to acquire new downstream skills with almost no forgetting.
We argue that these varied findings motivate moving beyond a model-level, task-agnostic perspective: plasticity and stability should be characterized at the level of individual task families and transformer modules, in relation to the functional roles each module plays in each task.

To systematically disentangle both tasks and architectural modules, cognitive science provides a foundational blueprint.
Decades of memory research establish a classic distinction between two fundamental systems: \textit{declarative} memory (``knowing that''), which supports factual, associative knowledge recall; and \textit{procedural} memory (``knowing how''), which governs multi-step rule execution and algorithmic skills \citep{cohen1980procedural,squire1992memory}.
Neuropsychological studies reveal a double dissociation between them, suggesting that declarative and procedural memory depend on at least partially distinct neural substrates: amnesic patients acquire procedural skills such as mirror reading despite having no recollection of the training episodes \citep{cohen1980procedural}, whereas patients with neostriatal dysfunction show the reverse pattern, failing to learn a habit-like classification task while still remembering the training episodes \citep{knowlton1996dissociation}.
Modern transformers exhibit an analogous distinction along both functional and structural dimensions.
Functionally, language models are simultaneously required to serve as vast factual encyclopedias (declarative tasks) and multi-step operational executors (procedural tasks).
Structurally, mechanistic interpretability research suggest an analogous functional specialization: feed-forward layers act as key-value stores for factual associations \citep{geva2021transformer,meng2022locating}, whereas attention circuits route information across positions and implements algorithmic computations \citep{elhage2021mathematical,olsson2022context}.
However, existing research examines this heterogeneity primarily under static conditions, without addressing how it affects the dynamic trade-off between plasticity and stability during continual adaptation.
Crucially, as our experiments show later, this relationship is not a simple binary mapping in which each module exclusively serves and affects one task family.

To rigorously analyze these dynamics, standard pre-trained models are too noisy: task boundaries are often ambiguous, prior data exposure is unknown, and the amount of memorized information is difficult to quantify independently of generalization.
To remove these confounds, we therefore design a clean synthetic testbed with two archetypal task families.
For declarative memory, we use random sequence memorization and quantify the amount of memorization in bits, following the information-theoretic framework of \citet{morris2026how}.
For procedural task, we utilize algorithmic graph traversal (BFS and DFS) evaluated by sequence-level accuracy.
Under a two-stage learning protocol, we evaluate both new-task plasticity and old-task stability across transformer modules.

Our synthetic experiments reveal a double dissociation: QK and VO projections are the most plastic on procedural tasks, whereas MLPs are the most plastic on declarative memorization.
More importantly, cross-task stability does not simply follow or reverse plasticity.
When learning a new procedure, updating QK retains over half of the stored declarative memory, whereas updating MLPs retains far less, and updating VO even loses roughly four-fifths, though VO and QK are equally plastic on the procedure itself.
We further probe the source of this asymmetry through targeted module reset-and-recovery experiments.
They show that procedural skill is primarily implemented in QK and VO, and declarative information is preferentially stored in MLPs, matching where each module is most plastic; yet QK and VO also support declarative memory by providing access to storage held in MLPs.
Updating a module can therefore impair a capability it does not primarily implement: VO, though a weaker storage site than MLPs, can disrupt declarative retention more severely.

Finally, we study a pretrained model (Qwen2.5-0.5B) on various realistic procedural and declarative tasks, showing that our findings largely carry over.
For instance, on the code execution task (procedural), QK-tuning achieves 29.4\% accuracy while retaining over 67.2\% of factual QA capability (declarative), whereas VO, MLP, and full tuning achieve similar or lower procedural performance but retain only 33.0\%, 6.8\%, 4.6\% declarative knowledge.
Together, these results indicate that plasticity and stability depend on different aspects of a module's functional role in a task, and therefore exhibit asymmetric patterns.
Our primary contributions are summarized as follows:

\begin{itemize}
    \item We frame continual learning at the level of task families and transformer modules, through the lens of cognitive memory systems (declarative vs.\ procedural) and their specialized neural substrates.
    \item We design a synthetic testbed that cleanly measures declarative memory in bits and procedural skill by exact execution, revealing a double dissociation in plasticity and an asymmetric, non-mirrored pattern in cross-task stability.
    \item Through targeted reset-and-recovery experiments, we provide empirical evidence that these properties depend on different aspects of a module's functional role: plasticity tracks a module's suitability for implementing a task's primary mechanics, whereas stability additionally reflects auxiliary roles such as access to storage held elsewhere, rather than a strict task-modular division.
    \item We show these findings largely carry over to fine-tuning a pretrained model on real-world procedural and declarative tasks, where modules reaching similar procedural performance can differ sharply in how much declarative knowledge they retain.
\end{itemize}

\section{Synthetic Testbed and Results}
\label{sec:synthetic}

To isolate the plasticity and stability dynamics from the confounds of pre-training data distributions, ambiguous task boundaries, and unquantifiable parametric memory, we build a testbed following the controlled synthetic approach of \cite{AllenZhu-icml2024-tutorial}. The testbed separates the memorization of arbitrary information from the execution of procedural rules, allowing us to measure declarative memory and procedural skill independently and quantitatively. In this section, we first introduce the two task families and their measurements, then describe the continual learning protocol used to conduct experiments, and finally analyze the observed learning and retention dynamics.


\begin{figure}[t]
    \centering
    \includegraphics[width=1.0\linewidth]{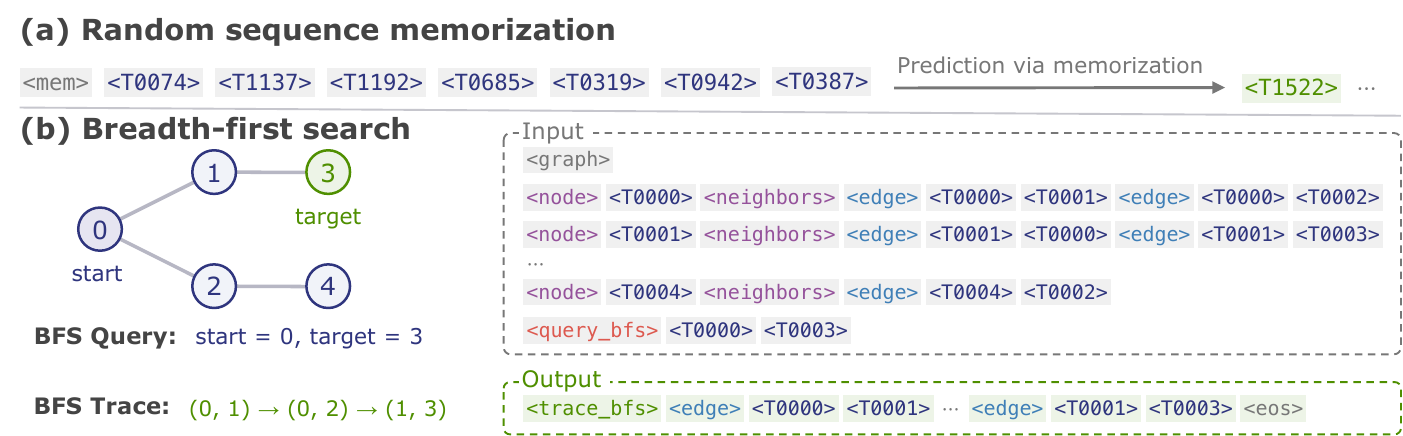}
    \caption{(a) Synthetic Declarative Task and (b) Procedural Task }
    \label{fig:syn_tasks}
\end{figure}
 
\subsection{Declarative and Procedural Tasks with Measurements}
\label{sec:tasks}

\paragraph{Declarative Task: Random Sequence Memorization}
To isolate information storage from the learning of generalizable rules, we define a fixed memory bank $\mathcal{D}_{\mathrm{mem}} = \{x^{(i)}\}_{i=1}^{N}$, where each sequence $x^{(i)}$ consists of $L$ tokens sampled independently and uniformly from a discrete vocabulary $\mathcal{V}$ of size $V$. We set $L = 64$ and $V = 2048$. The model learns these sequences through standard autoregressive next-token prediction. Because tokens are sampled independently, the generating distribution provides no structural or syntactic patterns that predict their continuation. Successful recall therefore depends on learning sequence-specific associations from the memory bank, making this task a controlled proxy for declarative memory.
Following \citet{morris2026how}, we quantify memorization as the reduction in description length relative to a uniform reference:\\
$
    \mathrm{mem}_U(\theta; \mathcal{D}_{\mathrm{mem}})
    = \sum_{x \in \mathcal{D}_{\mathrm{mem}}}
    \max\{ 0, (L-1)\log_2 V + \sum_{t=2}^{L} \log_2 p_\theta(x_t \mid x_{<t}) \}
$,
\\where predictions are conditioned on ground-truth prefixes, with the shared task prefix omitted from the notation. $\mathrm{mem}_U$ quantifies how much memorized declarative information about $\mathcal{D}_{\mathrm{mem}}$ can be read out from the current model $\theta$, measured in bits.
\Cref{fig:syn_tasks}(a) shows this task.

\paragraph{Procedural Task: Algorithmic Graph Traversal}
To test the execution of procedural tasks, we use breadth-first search (BFS) and depth-first search (DFS). For each task $\tau \in \{\mathrm{BFS}, \mathrm{DFS}\}$, we construct training and test sets $\mathcal{D}^{s}_{\tau} = \{((G_i, q_i), y_i)\}_{i=1}^{N^{s}_{\tau}}$, where $s \in \{\mathrm{train}, \mathrm{test}\}$.
As shown in \Cref{fig:syn_tasks}(b), each input provides a connected undirected graph $G_i$ and a traversal query $q_i$ in-context, and the model must generate the corresponding traversal trace $y_i = (e_{i,1}, \dots, e_{i,T_i})$ as a sequence of edges. Each graph contains 32 nodes with maximum degree 3, and thus the trace contains $T_i \in [2, 31]$ edges. For each task, we use $N^{\mathrm{train}}_{\tau} = 256\mathrm{k}$ training samples and $N^{\mathrm{test}}_{\tau} = 10\mathrm{k}$ test samples. The two tasks share the same graph representation and output format but differ in how they organize traversal, featuring two different execution procedures to learn. They require the model to track the evolving traversal state differently and select each subsequent step accordingly.
To assess procedural skill acquisition beyond memorizing training examples, we evaluate on held-out graphs. Our primary metric is sequence-level exact-match accuracy under autoregressive greedy decoding:
$\mathrm{Acc}_{\tau}(\theta; \mathcal{D}^{\mathrm{test}}_{\tau}) = \sfrac{ \sum_{((G,q),y) \in \mathcal{D}^{\mathrm{test}}_{\tau}} \mathbb{I}\left[\hat{y}_{\theta}(G, q) = y\right] }{N^{\mathrm{test}}_{\tau}}$.

The two tasks thus distinguish declarative memory of arbitrary observed content from procedural skill of executing on unseen inputs. Their respective measurements provide the basis for evaluating new-task learning and old-task retention under the continual learning protocol below.

\subsection{Continual Learning Protocol}
\label{sec:protocol}
 
Unlike training from random initialization on a single task, continual learning requires adapting a model that already possesses knowledge and skills. We therefore use a two-stage protocol to measure plasticity and stability from an established starting point, where new learning and its effects on prior capabilities can be observed together.
 
\paragraph{Stage 1: Establishing Declarative Memory and Procedural Skill}
We use the ``8L512D'' transformer (25M parameters) from \cite{Allenzhu2025-canon} in all synthetic experiments. Starting from random initialization, we first train the full model on $\mathcal{D}^{\mathrm{old}}_{\mathrm{mem}} \cup \mathcal{D}^{\mathrm{train}}_{\mathrm{BFS}}$ to establish both declarative memory and procedural skill. The resulting checkpoint, denoted by $\theta_1$, achieves $\mathrm{Acc}_{\mathrm{BFS}}(\theta_1; \mathcal{D}^{\mathrm{test}}_{\mathrm{BFS}}) = 93.25\%$ and a memory density of $\sfrac{\mathrm{mem}_U(\theta_1; \mathcal{D}^{\mathrm{old}}_{\mathrm{mem}})}{P} = 3.60$ bpp (bits per parameter), where $P$ is the total parameter count. This memory density matches the 3.6 bpp reported by \citet{morris2026how} because we set $|\mathcal{D}^{\mathrm{old}}_{\mathrm{mem}}| = 196\mathrm{k}$ based on their estimation to fully saturate the model, ensuring substantial memorization alongside procedural competence. We use this checkpoint as the common starting point for all Stage-2 adaptation experiments.
By default, all results are averaged over experiments with three random seeds.
 
\paragraph{Stage 2: New-Task Adaptation and Cross-Task Retention}
Starting from $\theta_1$, we train to obtain $\theta_2$ in two separate adaptation setups:
\begin{itemize}
    \item \textbf{Procedural Adaptation}: we train on $\mathcal{D}^{\mathrm{train}}_{\mathrm{DFS}}$, measuring plasticity by $\mathrm{Acc}_{\mathrm{DFS}}(\theta_2; \mathcal{D}^{\mathrm{test}}_{\mathrm{DFS}})$ and stability by the retention ratio $\sfrac{\mathrm{mem}_U(\theta_2; \mathcal{D}^{\mathrm{old}}_{\mathrm{mem}})}{\mathrm{mem}_U(\theta_1; \mathcal{D}^{\mathrm{old}}_{\mathrm{mem}})}$.
    \item \textbf{Declarative Adaptation}: we train on $\mathcal{D}^{\mathrm{new}}_{\mathrm{mem}}$ disjoint from $\mathcal{D}^{\mathrm{old}}_{\mathrm{mem}}$, measuring plasticity by $\mathrm{mem}_U(\theta_2; \mathcal{D}^{\mathrm{new}}_{\mathrm{mem}})$ divided by trainable parameter count in bpp and stability by the retention ratio $\sfrac{\mathrm{Acc}_{\mathrm{BFS}}(\theta_2; \mathcal{D}^{\mathrm{test}}_{\mathrm{BFS}})}{\mathrm{Acc}_{\mathrm{BFS}}(\theta_1; \mathcal{D}^{\mathrm{test}}_{\mathrm{BFS}})}$.
\end{itemize}
 
We deliberately focus on stability across task families: declarative memory after learning a new procedure, and procedural skill after learning new declarative information, to investigate whether learning for one type of capability interferes with a qualitatively different capability already present in the model. This axis is not directly captured by standard aggregate or within-task forgetting measurements in prior work, yet it is particularly informative for revealing that plasticity and stability follow different aspects of a module's functional role, as we show later.

\paragraph{Module-Wise Adaptation}
To study module-level dynamics, during Stage 2 we freeze all parameters except a targeted transformer module to update. We consider the query and key projections jointly as \textbf{QK}, the value and output projections jointly as \textbf{VO}, the complete self-attention block as \textbf{Attention}, and the feed-forward layers as \textbf{MLP}. We additionally report \textbf{Full} adaptation as a reference in which all model parameters are trainable. Because QK, VO have 4.2M parameters respectively, and MLP has 15.7M parameters, raw comparisons can otherwise confound architectural module type with trainable parameter count. We therefore include two trainable-parameter-matched MLP setups matching QK or VO parameter count: (1) \textbf{MLP-LoRA}: We apply rank-97 LoRA to all projections in every MLP layer, yielding 4.17M trainable parameters while freezing the original weights. (2) \textbf{MLP-S}: Since LoRA can produce distinct adaptation behavior, such as the emergence of intruder dimensions \citep{shuttleworth2025lora}, we also directly update a randomly sampled subset of MLP channels, yielding 4.19M trainable parameters.
 
\subsection{Results in Synthetic Testbed}
\label{sec:synthetic_results}

\newif\ifwrapmainfig
\wrapmainfigfalse   

\ifwrapmainfig

    \begin{wrapfigure}{r}{0.58\columnwidth}
        \centering

\else

    \begin{figure}[t]
        \centering
        \begin{minipage}{0.90\linewidth}
            \centering

\fi


\begin{subfigure}[t]{0.48\linewidth}
    \centering
    \includegraphics[width=\linewidth]{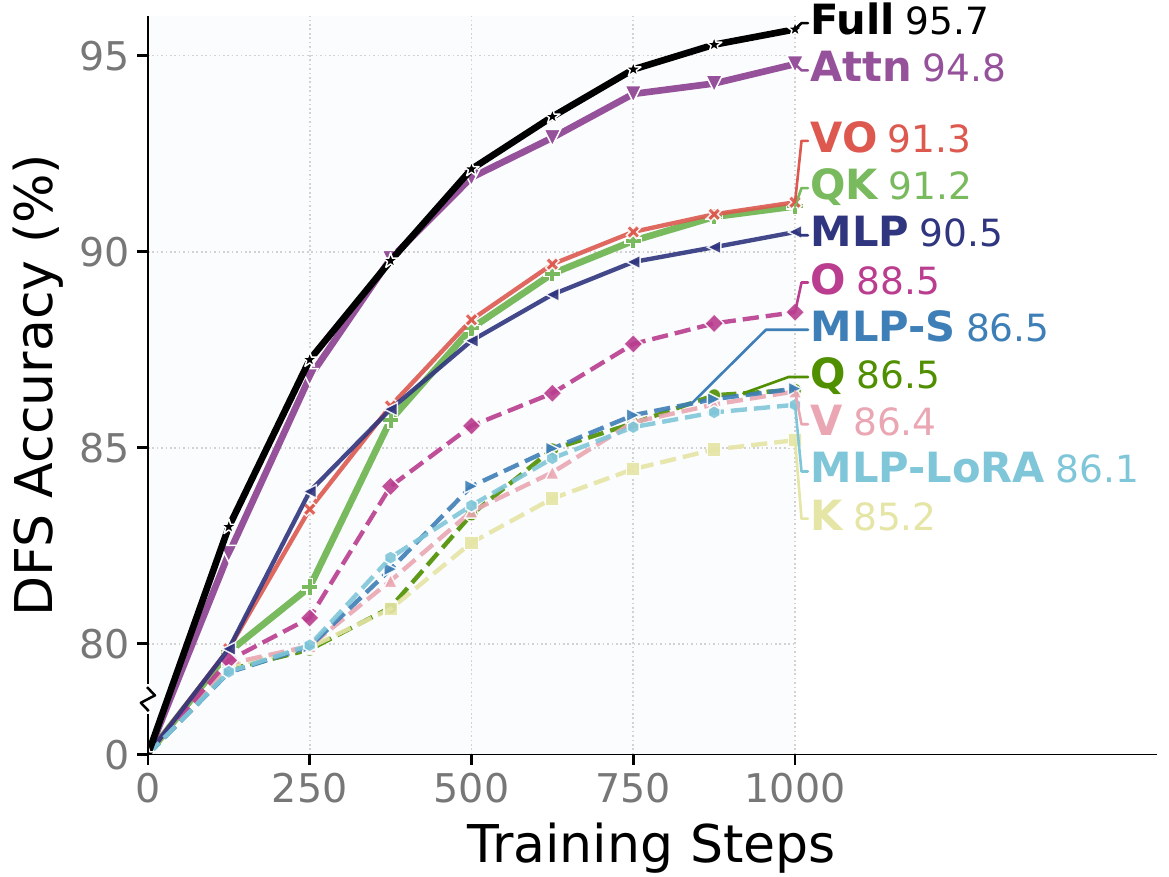}
    \caption{Plasticity of Procedural Adaptation}
    \label{fig:plasticity_dfs}
\end{subfigure}
\hfill
\begin{subfigure}[t]{0.48\linewidth}
    \centering
    \includegraphics[width=\linewidth]{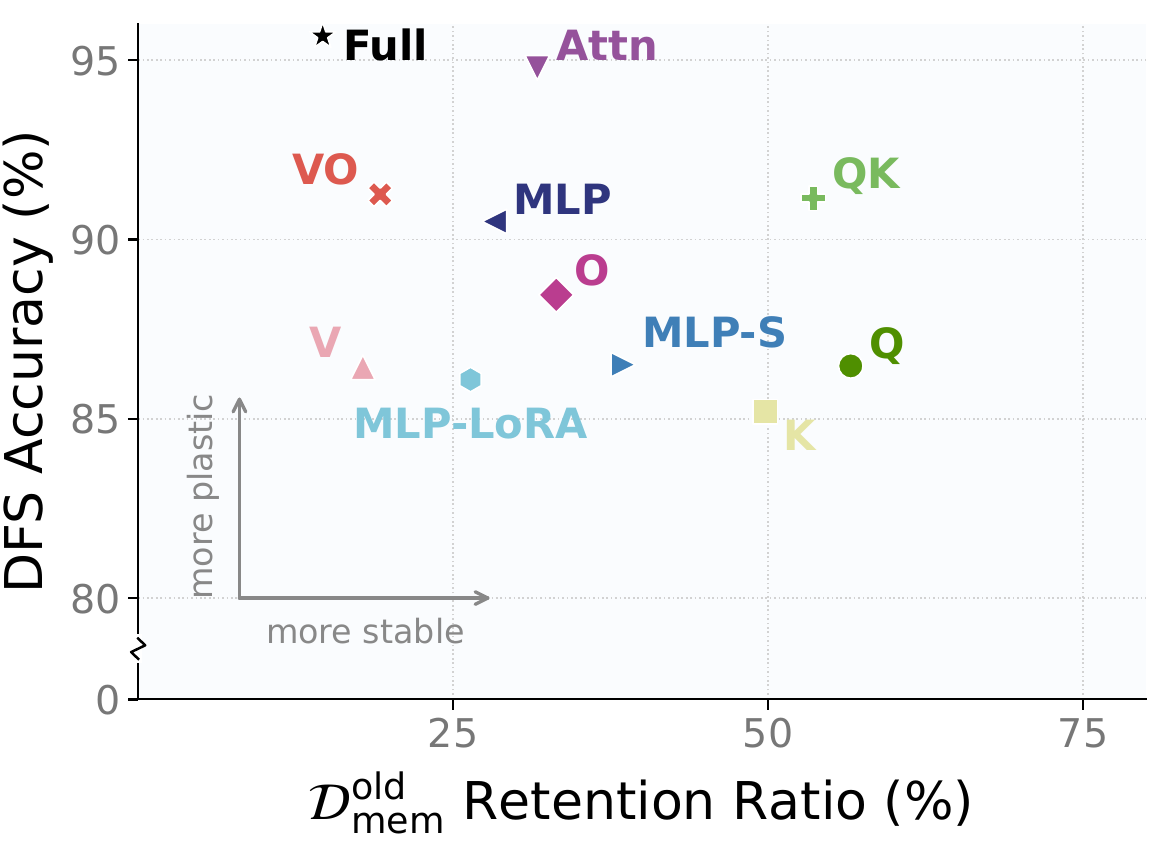}
    \caption{Trade-off in Procedural Adaptation}
    \label{fig:dfs_mem}
\end{subfigure}

\vspace{0.5em}

\begin{subfigure}[t]{0.48\linewidth}
    \centering
    \includegraphics[width=\linewidth]{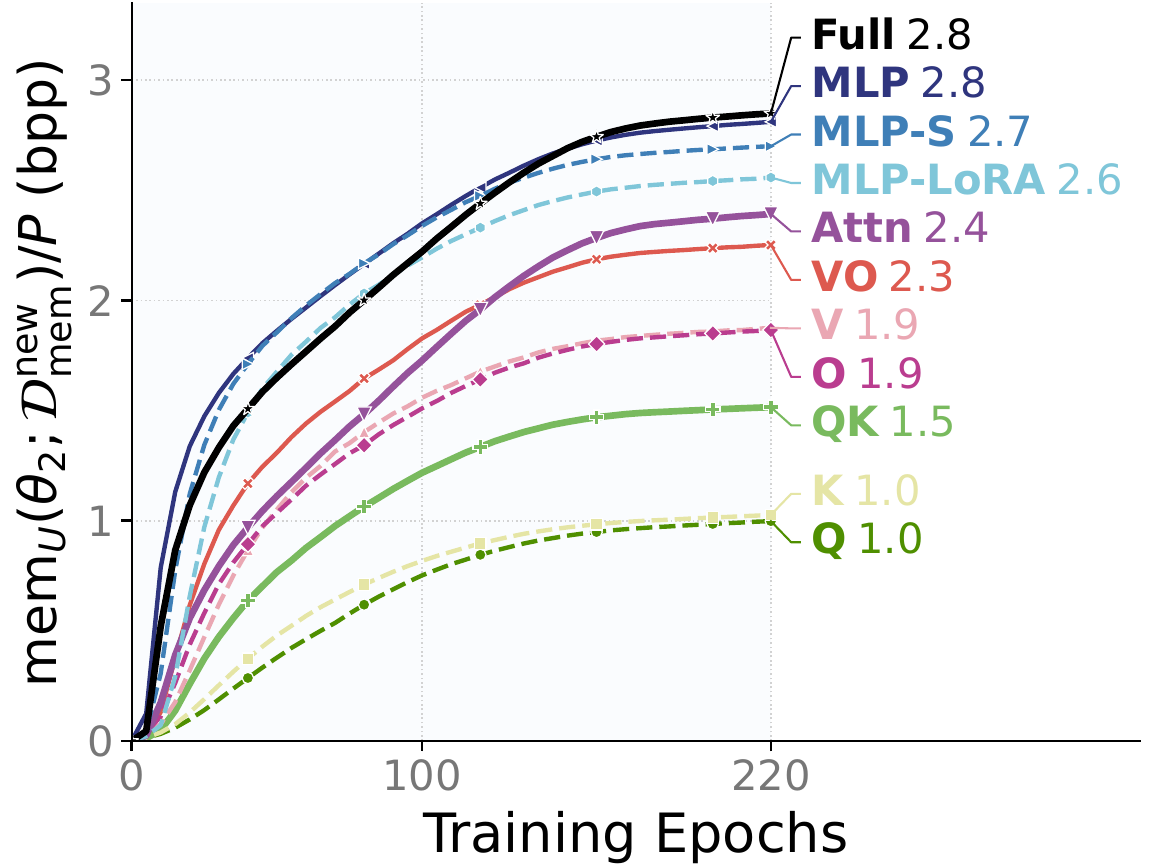}
    \caption{Plasticity of Declarative Adaptation}
    \label{fig:plasticity_mem}
\end{subfigure}
\hfill
\begin{subfigure}[t]{0.48\linewidth}
    \centering
    \includegraphics[width=\linewidth]{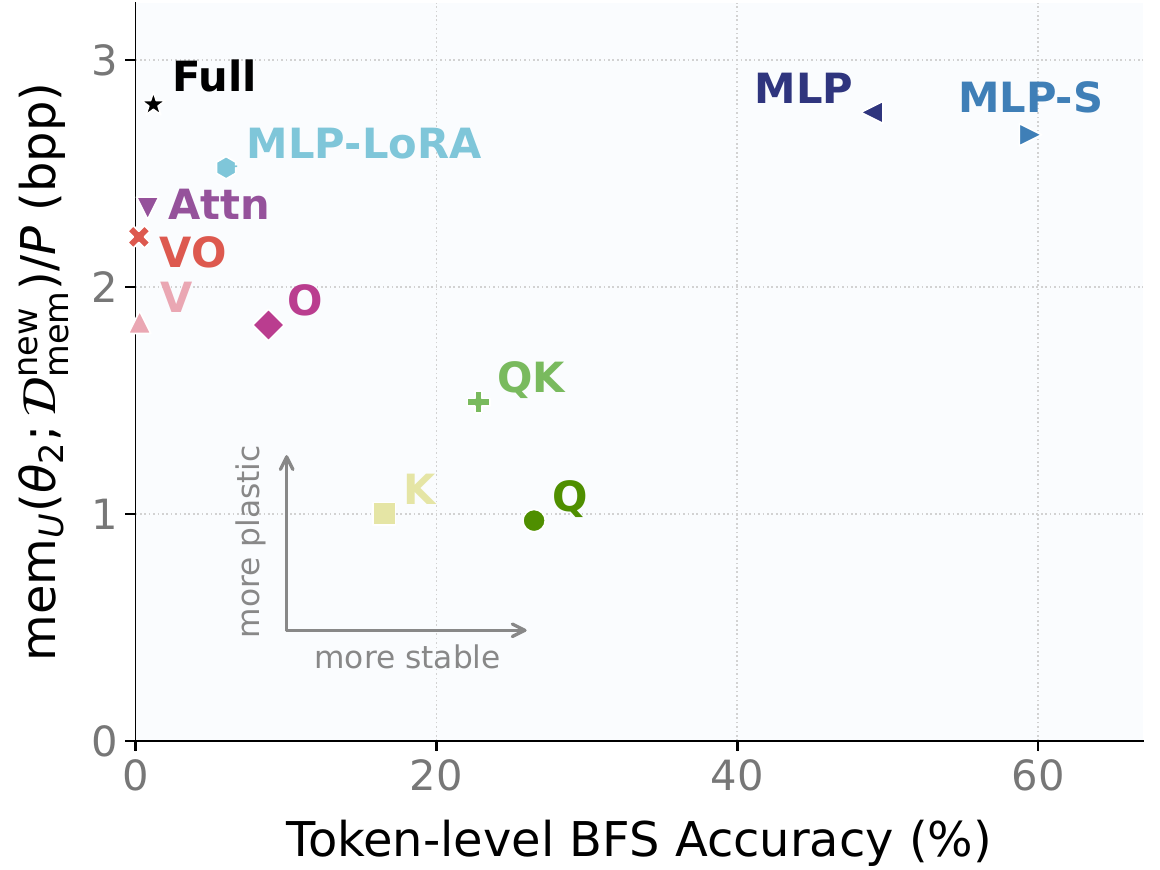}
    \caption{Trade-off in Declarative Adaptation}
    \label{fig:mem_bfs}
\end{subfigure}

\caption{Synthetic experiments show asymmetric plasticity and cross-task stability.}
\label{fig:syn_results}


\ifwrapmainfig

    \end{wrapfigure}

\else

        \end{minipage}
    \end{figure}

\fi

\paragraph{Plasticity Shows a Task--Module Double Dissociation}
We first compare how effectively different modules acquire a new capability from the common Stage-1 checkpoint.
As shown in \Cref{fig:plasticity_dfs}, when learning the new procedural task DFS, QK and VO learn faster, and achieve higher DFS accuracy (about 91.2\%) than MLP (90.5\%) and parameter-count-matched MLP setups (86.1-86.5\%). Conversely, on declarative adaptation (learning $\mathcal{D}^{\mathrm{new}}_{\mathrm{mem}}$), this plasticity order reverses. In \Cref{fig:plasticity_mem} comparing the capability of acquiring new declarative information, MLPs exhibit the highest plasticity on par with full-parameter learning \footnote{Even full model training cannot reach about 3.6 bpp anymore, possibly due to the loss of plasticity \citep{Dohare2024LossOP}}, reaching 2.80 bpp at full capacity, 2.70 bpp with MLP-S, and 2.60 bpp with MLP-LoRA, compared with 2.30 bpp for VO and 1.54 bpp for QK.
These results reveal a double dissociation in plasticity: QK and VO are more effective for learning a new procedure, whereas MLP is more effective for acquiring new declarative information.
Thus each task family favors different modules for acquiring a new capability.

\paragraph{Stability Exhibits a Distinct Task--Module Pattern}
We next compare how well previously acquired capabilities survive new-task learning. \Cref{fig:dfs_mem} shows that when learning the new procedure DFS, updating different modules exhibits markedly divergent stability of previously acquired declarative memory.
Updating QK is more stable, preserving about 54\% of the declarative information memorized in stage-1, compared with 28\% for MLP, 26\% for MLP-LoRA and 39\% for MLP-S.
By contrast, when learning new declarative information, the stability order of previously learned procedural skill is different. Under greedy autoregressive decoding, accumulating single-step prediction errors causes sequence-level accuracy to drop to zero across all module setups after declarative learning converges. However, inspecting teacher-forced token accuracy reveals differences in \Cref{fig:mem_bfs}: MLP-S and MLP retains 59.5\%, 49.0\% token accuracy, respectively, higher than 22.8\% for QK and almost zero for VO. 
Thus, stability also varies sharply across modules and task directions, but its pattern is distinct from the plasticity's above.

\paragraph{Cross-task Stability Does Not Mirror Plasticity}
A natural hypothesis is that stability is the flip side of plasticity: a module that absorbs the new task family easily, or that has little affinity for the old one, should leave the old capability intact.
However, our results above contradict this.
First, similar plasticity does not imply similar stability: VO and QK have nearly identical plasticity in learning DFS, yet updating VO retains only about 19\% of the stored declarative memory, far below the 54\% retained by QK.
Second, lower plasticity on a task family does not mean higher stability on the same family: VO stores new declarative information less effectively than MLPs (2.30 vs.\ 2.60--2.80 bpp), yet updating VO for DFS erases more of the old declarative memory than MLPs (19\% vs.\ 26--39\%).
Together, these results show that plasticity and cross-task stability cannot be reduced to a single mirrored module ordering. This asymmetry motivates the targeted interventions in the next section, where we examine which aspects of a module's functional role are associated with each property.

\section{Functional Attribution via Targeted Reset-and-Recovery}
\label{sec:reset_recovery}

The asymmetries above suggest that plasticity and stability may follow different aspects of how a module contributes to a capability.
However, attributing a capability to an individual module in a naturally full-parameter-trained transformer is nontrivial. Removing any single module destroys both capabilities, every module can implement either task family to some extent given sufficient training, and $\mathrm{mem}_U$ cannot be decomposed into per-module storage.
We therefore use an intervention that asks how central a module is to a task family, rather than whether it is necessary. We reinitialize one module of a trained model each time, retrain the full model, and track how the task capability recovers.
If the reset module plays an auxiliary role, the core implementation survives in the remaining modules, and thus the capability will be restored quickly. If the module plays a primary role, the capability must be rebuilt, or the information it held is lost.
As we show below, this intervention separates a module's primary role in a capability from supporting roles.


\begin{figure}[t]
    \centering
    \begin{minipage}{1.0\linewidth}
        \centering
        \begin{subfigure}[t]{0.32\linewidth}
            \centering
            \includegraphics[width=\linewidth]{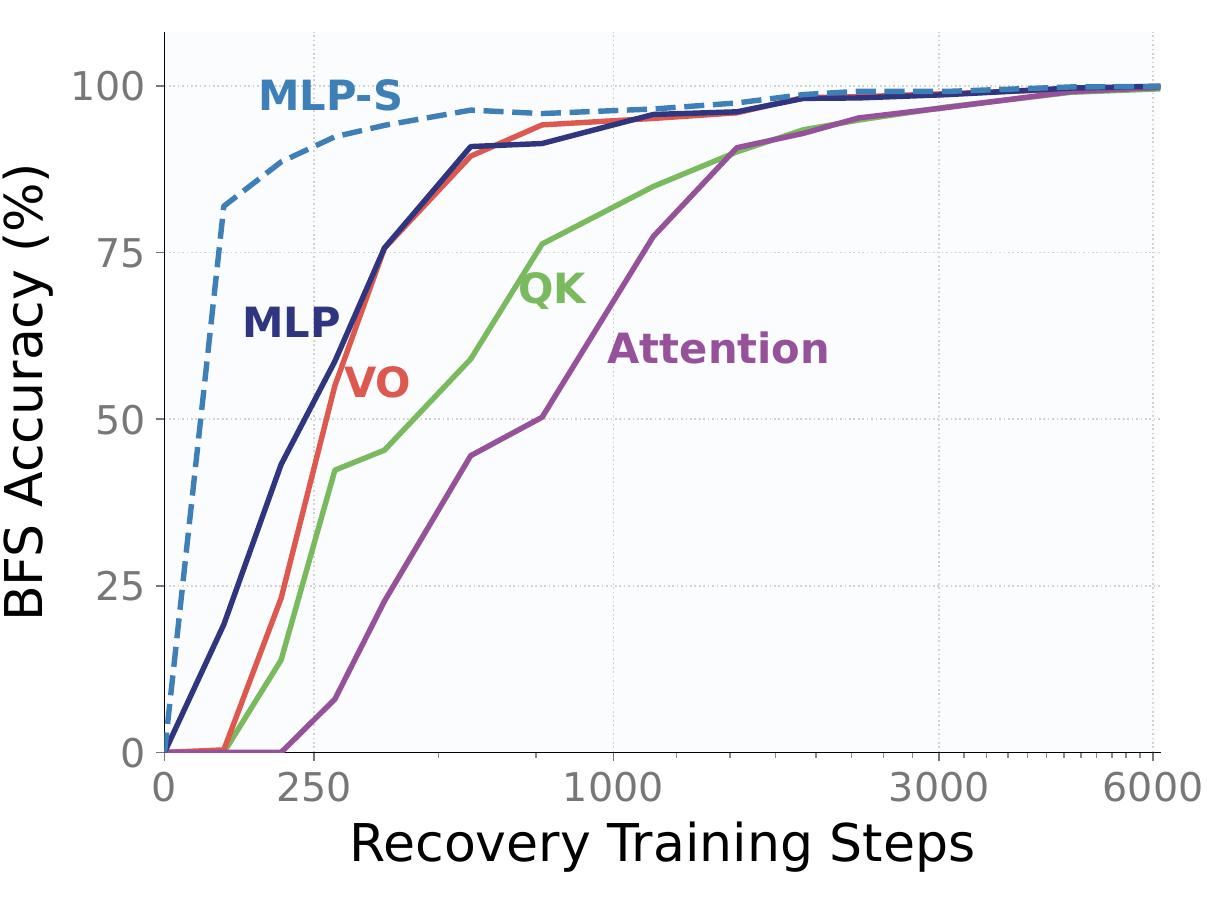}
            \caption{BFS Recovery Process}
            \label{fig:reset_bfs}
        \end{subfigure}\hfill
        \begin{subfigure}[t]{0.32\linewidth}
            \centering
            \includegraphics[width=\linewidth]{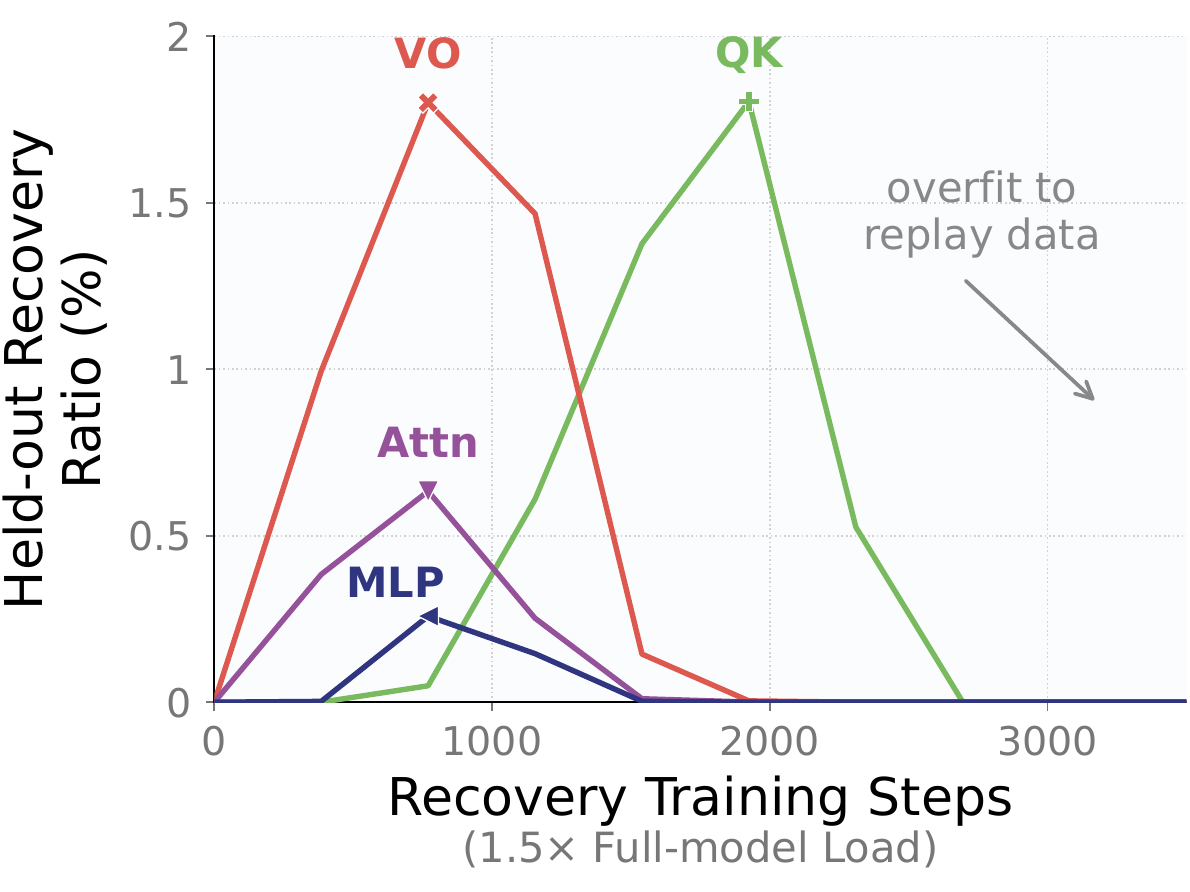}
            \caption{Memory Recovery Process}
            \label{fig:recovery_mem_full_load}
        \end{subfigure}\hfill
        \begin{subfigure}[t]{0.32\linewidth}
            \centering
            \includegraphics[width=\linewidth]{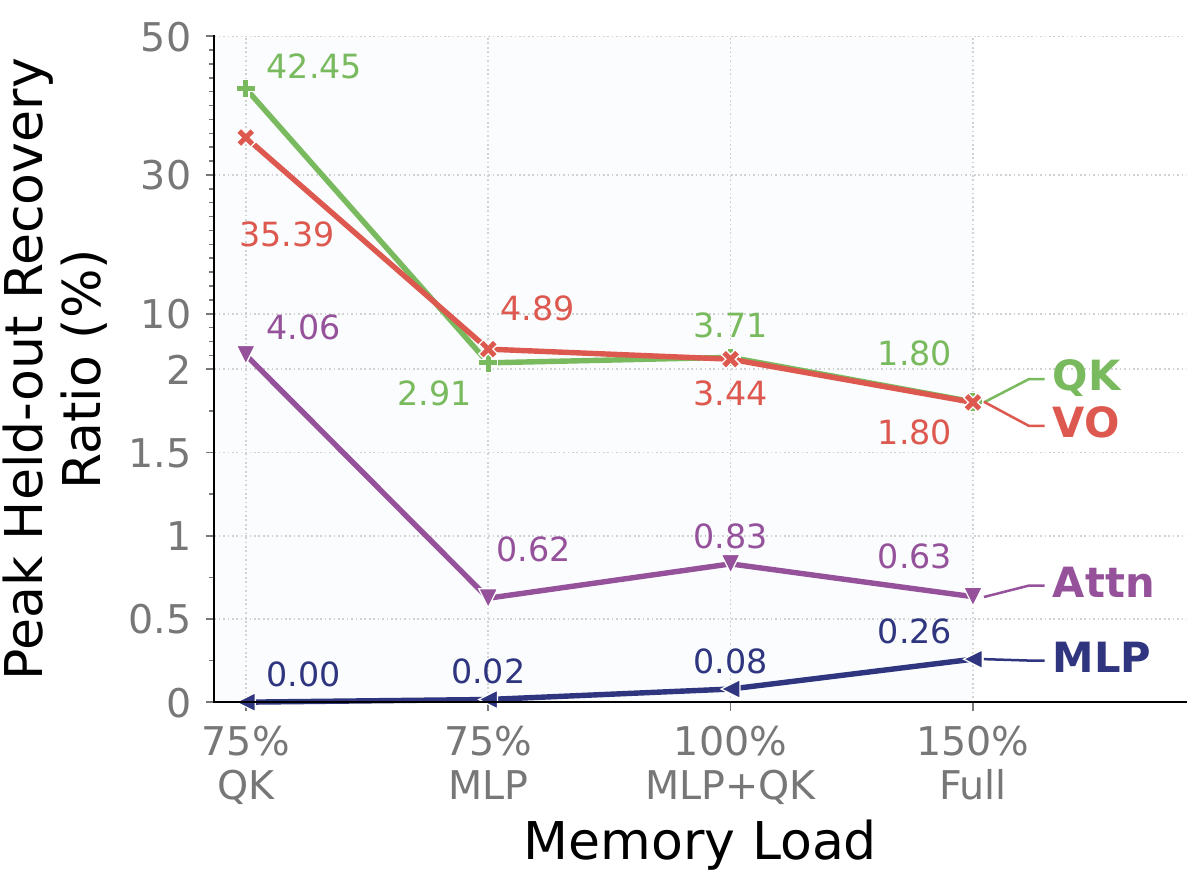}
            \caption{All-load Memory Recovery}
            \label{fig:recovery_mem}
        \end{subfigure}
    \end{minipage}
    \caption{Module-wise Reset-and-Recovery on Procedural Skill and Declarative memory}
    \label{fig:recovery}
\end{figure}

\subsection{Procedural Skill Is Primarily Implemented in QK and VO}

Starting from $\theta_1$, we reset QK, VO, Attention, or MLP and retrain the full model on $\mathcal{D}^{\mathrm{train}}_{\mathrm{BFS}}$. All setups eventually return to near-perfect BFS accuracy, so we compare early recovery efficiency rather than the final plateau.
As shown in \Cref{fig:reset_bfs}, resetting MLP recovers fastest, resetting VO is next, resetting QK is more slowly, and resetting the entire attention block is slowest.
Resetting a parameter-matched random subset of MLP channels leaves BFS largely intact and recovers the fastest, confirming that the gap is not explained by the number of reset parameters.
Also, MLPs are not simply faster procedural learners themselves, as \Cref{sec:synthetic_results} shows MLPs learn procedural tasks less efficiently than Attention, QK and VO.
Together, these observations suggest that the fast recovery after an MLP reset reflects procedural skills surviving in the remaining modules, whereas resetting QK or VO removes more of the learnt procedure, which must then be rebuilt, thus leading to slower convergence.
This indicates that procedural skill is primarily implemented in the attention projections, whereas MLPs contribute in a supporting role that is easier to restore, matching where procedural plasticity is highest in \Cref{sec:synthetic_results}.

\subsection{Declarative Memory Relies on MLP Storage and Attention Access}

To probe declarative memory, we start from a checkpoint $\theta_m$ sufficiently trained only on $\mathcal{D}_{\mathrm{mem}}$, reset one module, and retrain the full model on a 10\% subset of $\mathcal{D}_{\mathrm{mem}}$ as replay to obtain $\theta_m^r$. We then measure $\mathrm{mem}_U$ on the remaining 90\% that is never replayed.
Unlike procedural skill, arbitrary sequences cannot be relearned without re-exposure, so held-out information can reappear only if it survived in the remaining modules, with replay merely restoring the auxiliary function (e.g., access or readout existing storage) of the reset module. If the reset module itself held some information, the corresponding $\mathrm{mem}_U$ should stay near zero.
Because every module can store declarative information (as shown in \Cref{fig:plasticity_mem}) to some extent, we vary the memory load based on the 3.6 bpp estimation confirmed in \Cref{sec:protocol}: 75\% of the maximum capacity of QK (so theoretically QK or VO alone can fully store), 75\% of that of MLP (theoretically MLP can fully store, whereas QK+VO cannot), 100\% of the combined MLP+QK capacity (where MLP alone no longer suffices), and $1.5\times$ that of the full model (the same as $\mathcal{D}_{\mathrm{mem}}^{\mathrm{old}}$, where no standalone module can fully store).
We report the peak recovered ratio $\sfrac{\mathrm{mem}_U(\theta_m^r; \mathcal{D}^{\mathrm{held-out}}_{\mathrm{mem}})}{\mathrm{mem}_U(\theta_m; \mathcal{D}^{\mathrm{held-out}}_{\mathrm{mem}})}$, since prolonged training eventually overfits the replay subset and erases all held-out information, as shown in \Cref{fig:recovery_mem_full_load}.

As shown in \Cref{fig:recovery_mem}, the four loads reveal distinct contributions of storage and access to declarative memory.
Under the two lighter loads, resetting MLP leaves essentially no recoverable held-out memory (0.00\% and 0.02\%).
This holds even under the 75\% QK load, where QK or VO alone has sufficient capacity to hold the entire memory bank, potentially allowing MLP reset-and-recovery to read it out.
By contrast, resetting QK or VO leaves part of the held-out memory recoverable: 42.45\% and 35.39\% under the 75\% QK load, and 2.91\% and 4.89\% under the 75\% MLP load, respectively.
Together, these contrasting results suggest that MLPs are the preferred storage site for declarative information when their capacity suffices, whereas QK and VO serve mainly to \emph{access} the storage.
Limited replay cannot restore the information lost with the reset module but can partially restore the access to information, allowing information preserved in the remaining modules to become readable again.
This echoes prior findings that sparse replay can mitigate forgetting in language models \citep{demasson2019episodic, scialom-etal-2022-fine, zheng2025spurious}.

Recovery after resetting QK or VO declines as the load grows, falling from 35-42\% to 3--5\% and then to about 1.8\% under the $1.5\times$ full-model load, likely because access to a larger declarative memory is more difficult to recover through replay.
Against this trend, recovery after resetting MLP increases once the memory load exceeds MLP capacity: it is nearly zero under the two lighter loads, becomes small but positive (0.08\%) at the 100\% MLP+QK load, and reaches 0.26\% under the highest load.
These opposite trends suggest that the increase under MLP-reset-and-recovery is not simply due to heavier loads making reset-and-recovery easier.
Instead, they indicate that once the memory load exceeds what MLPs can hold, QK and VO also take on part of the storage, and that part can survive an MLP reset.
Thus, storage and access are not an exclusive assignment of roles to modules, but different aspects of a module's functional contributions, and declarative memory depends on both.

\subsection{Connecting Functional Roles to Plasticity and Stability}
\label{sec:connect}

These interventions suggest that plasticity and stability relate to different aspects of a module's functional role (\Cref{sec:synthetic_results}).
Plasticity aligns with which module is best suited to implementing a task's primary mechanics. Attention projections, which primarily implement procedural execution, are the most plastic in procedural learning. MLPs, the preferred storage site for declarative information, are the most plastic in declarative learning.
Stability, however, reflects the broader set of roles a module plays in an existing capability, including supporting roles beyond where the capability is preferentially implemented. Updating a module can therefore interfere not only with information directly stored there, but also with the access to information stored elsewhere.
In particular, VO's poor declarative retention under procedural adaptation is consistent with disrupting memory access, despite its weaker storage contribution than MLPs.
Its lower declarative retention than QK may reflect its more critical role in memory access, or greater vulnerability of that role to our specific procedural learning updates.
This interpretation is further supported by mixing only 1\% of $\mathcal{D}_{\mathrm{mem}}^{\mathrm{old}}$ into Stage-2 DFS procedural learning: while maintaining similar DFS performance, VO now retains 75.52\% of the declarative memory, close to QK (80.14\%) and well above MLP (56.34\%). This reverses the no-replay ordering between VO and MLP (19\% vs.\ 28\%), supporting that VO mainly serves the access role, which is easier to preserve or restore than MLP's storage.
Together, these findings indicate that plasticity tracks a module's suitability for implementing a task's primary mechanics, whereas stability reflects the full set of functional dependencies an existing capability has on the module, including supporting roles. Because where a capability is preferentially implemented does not fully capture these dependencies, plasticity and stability exhibit asymmetric rather than mirrored patterns.


\begin{figure}[h]
  \centering
  \captionsetup[subfigure]{font=scriptsize,skip=2pt}
  \begin{subfigure}[t]{0.24\linewidth}
    \centering
    \includegraphics[width=\linewidth]{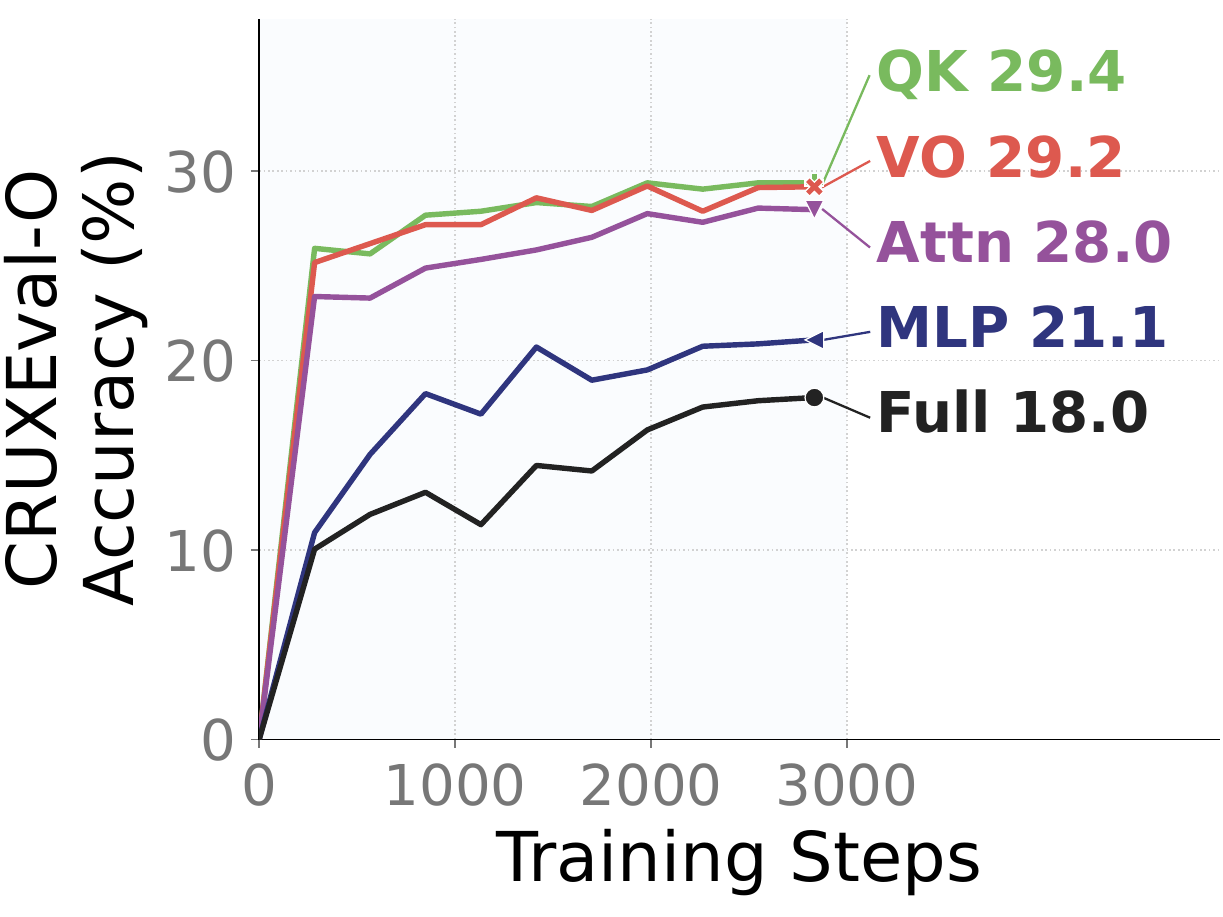}
    \caption{Plasticity on Code Execution}
    \label{fig:crux_learning_curve}
  \end{subfigure}\hfill
  \begin{subfigure}[t]{0.24\linewidth}
    \centering
    \includegraphics[width=\linewidth]{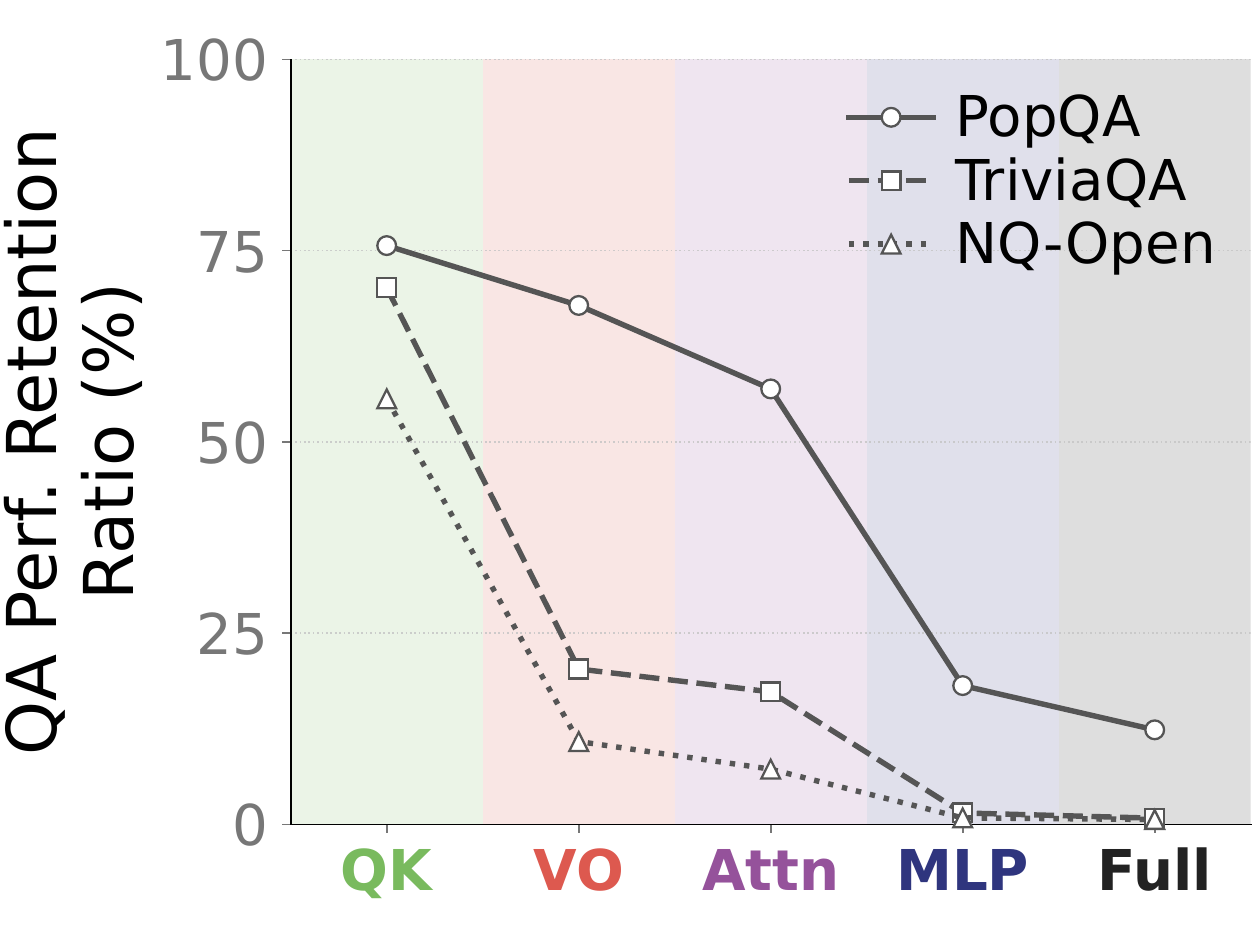}
    \caption{QA Stability in \ref{fig:crux_learning_curve}}
    \label{fig:crux_knowledge_retention}
  \end{subfigure}\hfill
  \begin{subfigure}[t]{0.24\linewidth}
    \centering
    \includegraphics[width=\linewidth]{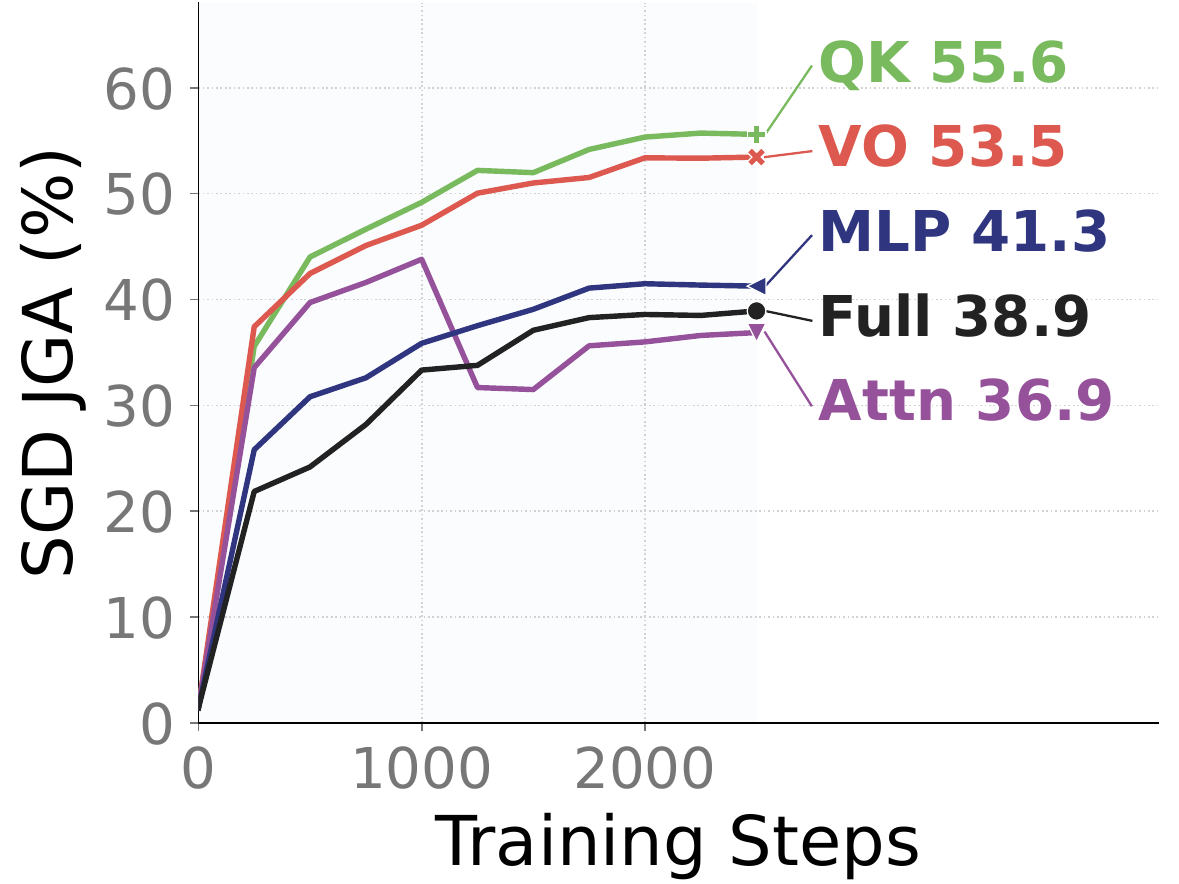}
    \caption{Plasticity on SGD}
    \label{fig:sgd_learning_curve}
  \end{subfigure}\hfill
  \begin{subfigure}[t]{0.24\linewidth}
    \centering
    \includegraphics[width=\linewidth]{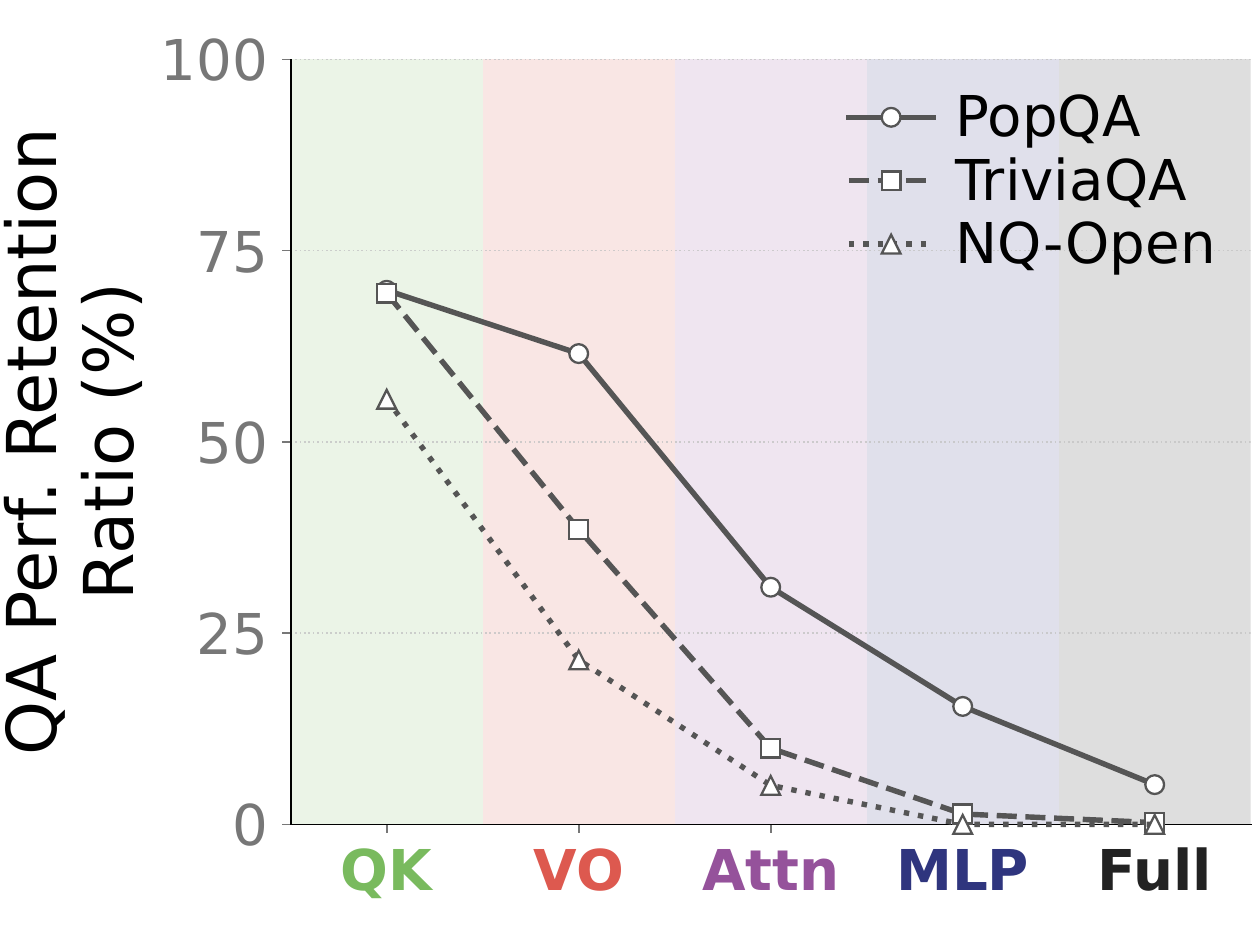}
    \caption{QA Stability in \ref{fig:sgd_learning_curve}}
    \label{fig:sgd_knowledge_retention}
  \end{subfigure}\par\medskip
  \begin{subfigure}[t]{0.24\linewidth}
    \centering
    \includegraphics[width=\linewidth]{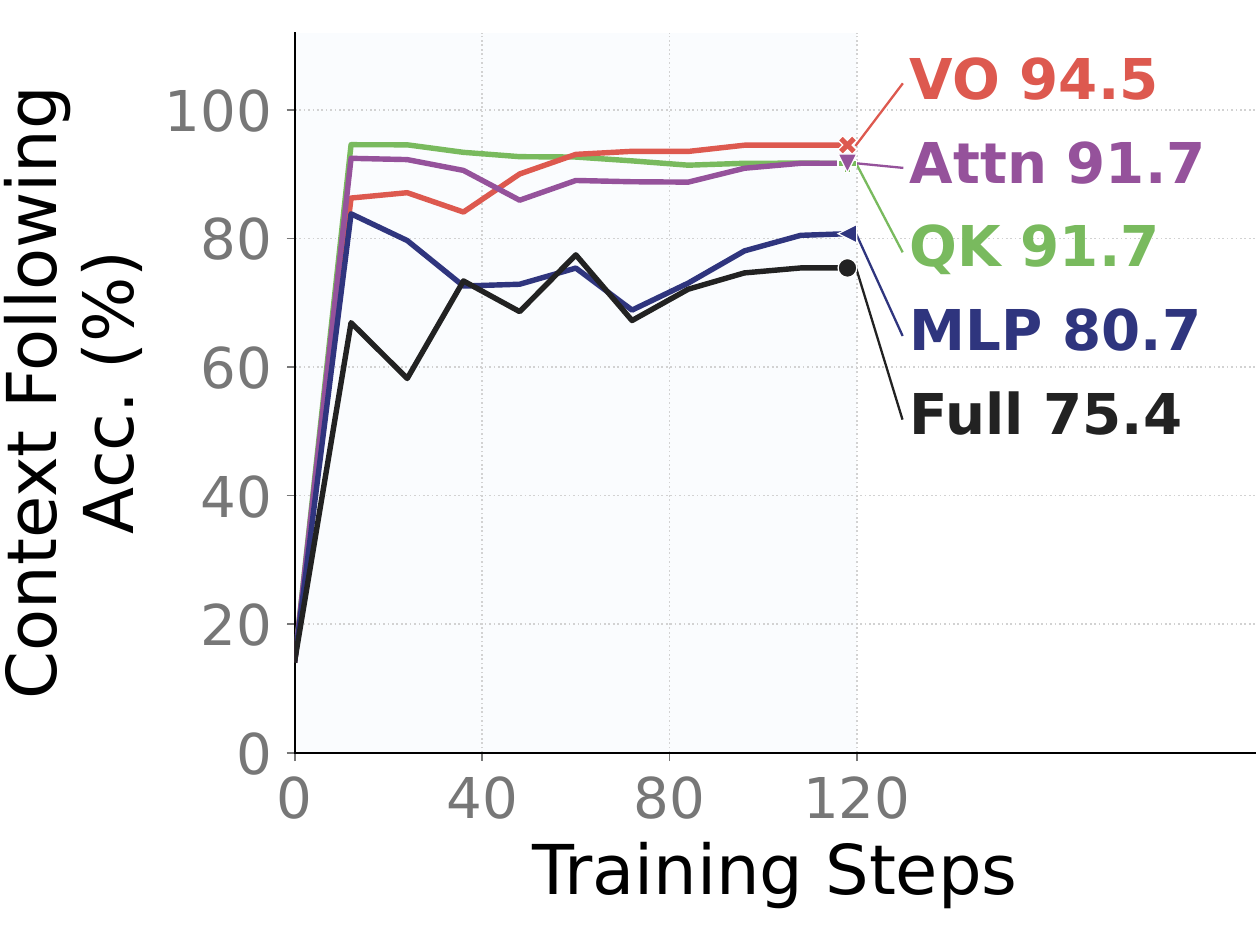}
    \caption{Plasticity on Ctx. Following}
    \label{fig:ctx_learning_curve}
  \end{subfigure}\hfill
  \begin{subfigure}[t]{0.24\linewidth}
    \centering
    \includegraphics[width=\linewidth]{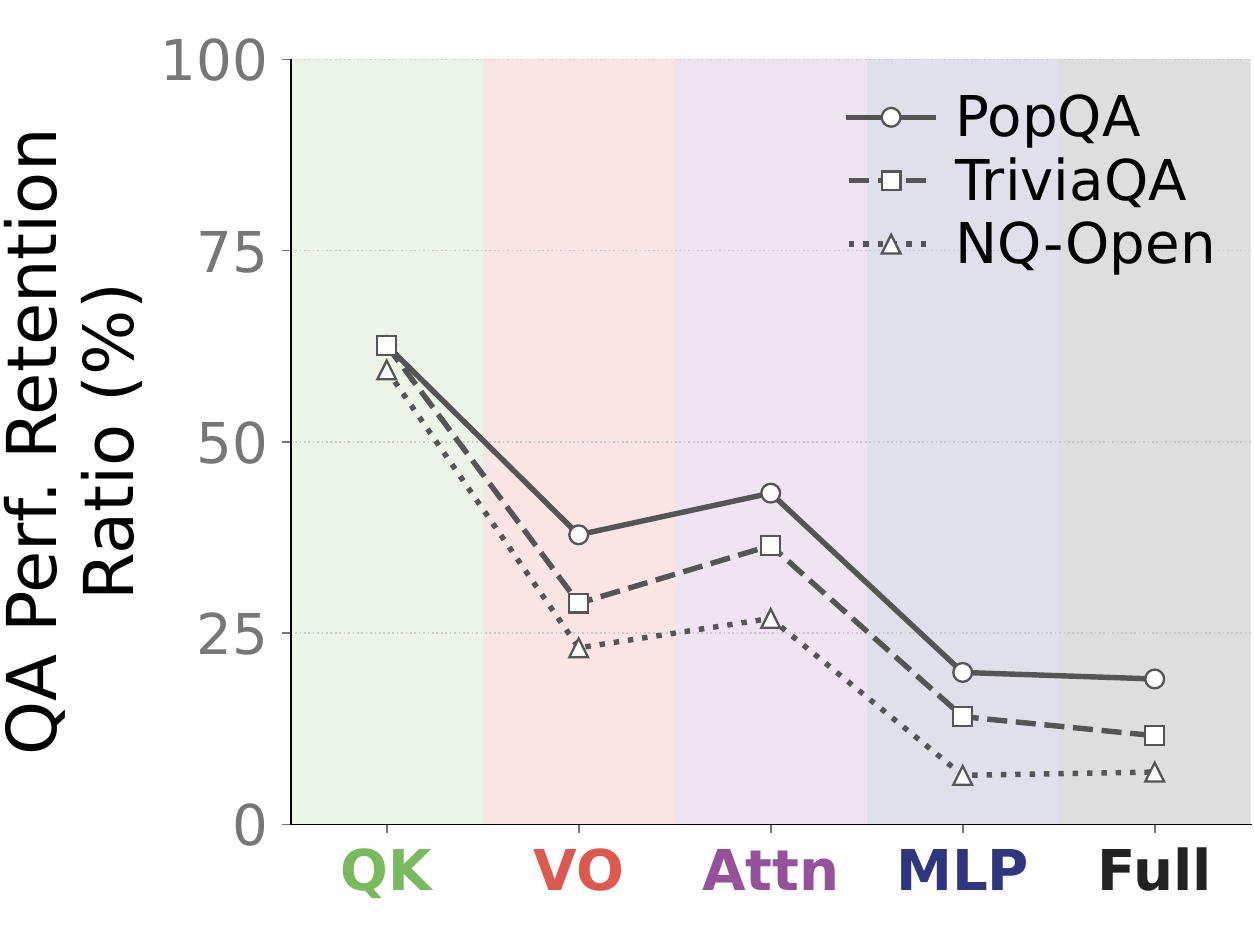}
    \caption{QA Stability in \ref{fig:ctx_learning_curve}}
    \label{fig:ctx_knowledge_retention}
  \end{subfigure}\hfill
  \begin{subfigure}[t]{0.24\linewidth}
    \centering
    \includegraphics[width=\linewidth]{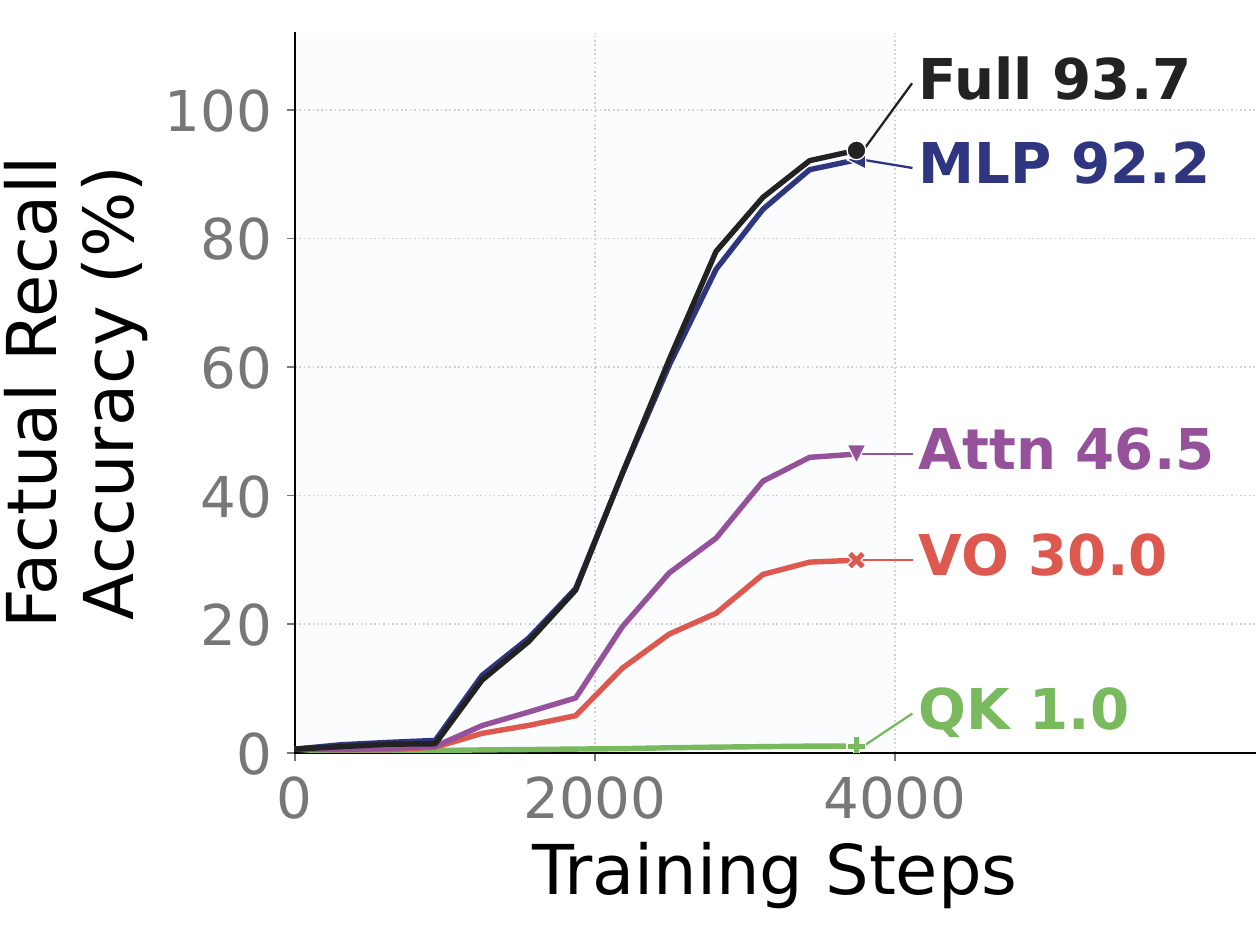}
    \caption{Plasticity on QA}
    \label{fig:qa_learning_curve}
  \end{subfigure}\hfill
  \begin{subfigure}[t]{0.24\linewidth}
    \centering
    \includegraphics[width=\linewidth]{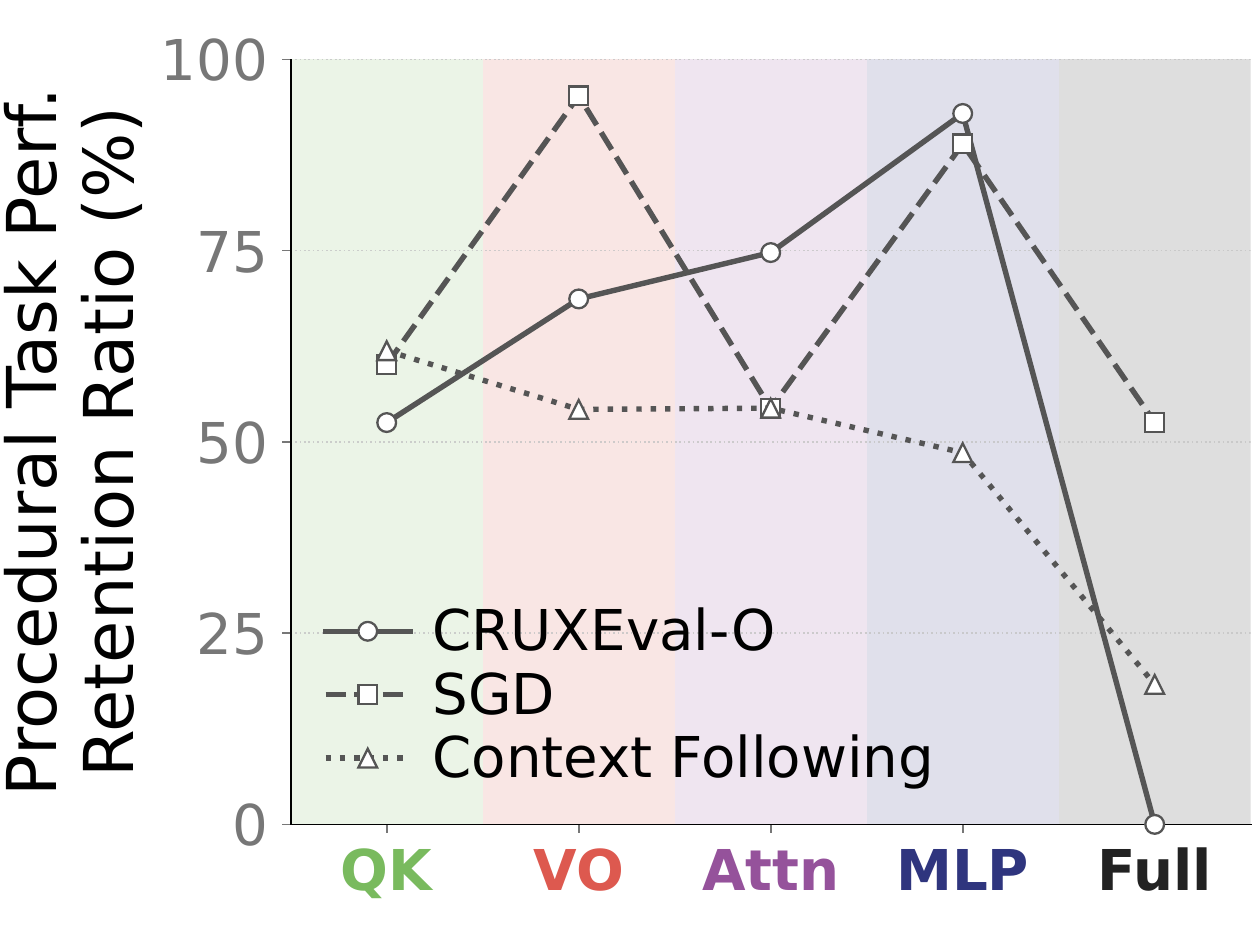}
    \caption{Procedural Stability in \ref{fig:qa_learning_curve}}
    \label{fig:qa_procedural_retention}
  \end{subfigure}
  \caption{Learning Curves and Respective Cross-Task Retention Levels on Realistic Tasks}
  \label{fig:real-results}
\end{figure}

\section{Realistic Setups and Results}
\label{sec:real}

We now ask whether the same picture holds when a pretrained large language model is updated on realistic tasks.
We use Qwen2.5-0.5B \citep{qwen2.5} as the base model $\theta_{\mathrm{p}}$ and, as in Stage 2 of \Cref{sec:protocol}, update a single module each time (QK, VO, Attention, or MLP) while freezing the rest, with Full as a reference, to obtain $\theta_{\text{ft}}$.
Since our goal is to test whether the findings in previous sections transfer rather than to maximize continual learning performance, we deliberately choose a relatively weak model that leaves substantial headroom across tasks to observe plasticity, instead of recent stronger models that already saturate many of them.

\paragraph{Realistic Tasks}
We select three procedural tasks that require input-dependent operations whose outcomes cannot be recalled from in-weight knowledge:
(1) \textbf{Code Execution}: given an in-context program and input, predict the output by simulating execution; we train on \textsc{PyX} \citep{peng2024semcoder} and evaluate output accuracy on \textsc{CRUXEval-O} \citep{gu2024cruxevalbenchmarkcodereasoning}.
(2) \textbf{Dialogue-to-API Conversion}: given a user-assistant dialogue and an API schema, generate the API call reflecting the latest dialogue state; we train and evaluate on Schema Guided Dialogue (SGD) \citep{Lee_2022}, and measure fuzzy joint goal accuracy (JGA).
(3) \textbf{Contextual Information Following}: given a passage and a question, answer strictly according to the passage even when it conflicts with in-weight knowledge; we construct splits from the three QA sources below and measure strict-following accuracy.
For declarative tasks, we use factual question answering focused on recall rather than complex reasoning\footnote{Reasoning tasks such as math can be viewed as a composite of both families: the model must recall axioms, theorems, and learned solution strategies (declarative), then apply input-specific operations to the given problem (procedural).}, sampling 5k QA pairs each from PopQA \citep{mallen-etal-2023-trust}, NQ-Open \citep{lee-etal-2019-latent}, and TriviaQA \citep{JoshiTriviaQA2017}.
\Cref{fig:real_tasks} shows these tasks.

\paragraph{Adaptation and Evaluation Protocol}
For procedural adaptation, we fine-tune $\theta_{\mathrm{p}}$ separately on each procedural task, evaluate plasticity by the task's accuracy metric, and measure stability as the retention ratio on factual QA, i.e., the proportion of questions correctly answered by $\theta_{\mathrm{p}}$ that remain correct after adaptation.
For declarative adaptation, because $\theta_{\mathrm{p}}$ has near-zero procedural accuracy, we first establish all three procedural skills through mixed training and then fine-tune on the 15k pooled QA pairs.
Plasticity is strict closed-book factual recall on the trained QA pairs, measuring memorization rather than held-out generalization, and stability is the ratio of each procedural task's performance after to before QA learning.

\paragraph{Plasticity Preferences Carry Over}
The top row of \Cref{fig:real-results} reproduces the double dissociation.
On all three procedural tasks, attention projections learn faster and reach higher accuracy than MLP: QK and VO reach 29.4\% and 29.2\% on \textsc{CRUXEval-O} versus 21.1\% for MLP (\Cref{fig:crux_learning_curve}), 55.6\% and 53.5\% JGA on SGD versus 41.3\% (\Cref{fig:sgd_learning_curve}), and 91.7\% and 94.5\% context-following accuracy versus 80.7\% (\Cref{fig:ctx_learning_curve}).
The preference reverses for factual acquisition (\Cref{fig:qa_learning_curve}): MLP reaches 92.2\% recall, on par with Full (93.7\%), whereas Attention, VO, and QK reach only 46.5\%, 30.0\%, and 1.0\%.
The dissociation is thus even sharper than in the synthetic testbed, with QK nearly unable to effectively memorize new facts.

\paragraph{Cross-Task Stability Remains Asymmetric}
As in \Cref{sec:synthetic_results}, similar plasticity does not imply similar stability.
Although QK and VO reach close procedural accuracy (\Cref{fig:crux_learning_curve,fig:sgd_learning_curve,fig:ctx_learning_curve}),
the retention panels (\Cref{fig:crux_knowledge_retention,fig:sgd_knowledge_retention,fig:ctx_knowledge_retention}) demonstrate that QK retains far more factual QA than VO (e.g., 70.2\% vs.\ 20.3\% on TriviaQA after learning code execution), and the QA stability ordering QK $>$ VO $>$ MLP $>$ Full holds across all procedural tasks learning.
This matches the access-disruption signature in \Cref{sec:reset_recovery}: VO is not the primary declarative storage site, yet updating it disrupts declarative knowledge far more than updating QK.
Unlike in the synthetic testbed, updating MLP here loses more declarative knowledge than updating VO.
A plausible reason is that realistic procedural tasks are not purely procedural, still drawing on in-weight knowledge, so procedural training may contain implicit declarative replay that preserves VO's access role, whereas facts overwritten in MLP storage cannot be restored this way.
This matches the synthetic setting with 1\% declarative replay, where VO retains more than MLP (\Cref{sec:connect}).
In the reverse direction (\Cref{fig:qa_procedural_retention}), updating MLP lets the model acquire far more trained facts while retaining most procedural task performance, echoing the synthetic finding that MLP updates best preserve procedural skill that is implemented mainly in attention projections (\Cref{sec:reset_recovery}).
Context following is the exception, plausibly because it requires answering from the passage even when it contradicts the model's own knowledge, and newly memorized facts can strengthen exactly this competing knowledge.

\section{Limitations and Future Work}

Our work provides a first view of plasticity and stability through module function, while both the architectural and functional dimensions can be studied at a finer resolution.

\paragraph{Finer architectural granularity.}
We treat QK, VO, and MLP as units of analysis, although each spans many layers and heads that may play different roles. Prior work already reveals submodule specialization: induction behavior is associated with particular attention heads and depths \citep{olsson2022context}, MLP functions vary across layers \citep{geva2021transformer}, and factual recall is mediated strongly by middle-layer MLPs \citep{meng2022locating}.
These findings suggest that the module-level roles we observe may themselves be distributed unevenly across finer components. Extending our analysis to individual layers, heads, and circuits could reveal whether storage, access, and procedural computation are concentrated in distinct subcomponents, and whether these subcomponents exhibit different plasticity--stability dynamics.
This could support more selective adaptation that updates the components most useful for a new task while preserving those important to existing capabilities.

\paragraph{Finer functional granularity.}
We use declarative and procedural memory as archetypal functional demands rather than a complete quantitative taxonomy. Realistic tasks may rely on both to different degrees, motivating a characterization of task composition along at least two continuous dimensions: the amount of stored information that must be retrieved and the amount of input-conditioned computation that must be performed.
In our synthetic setting, declarative load can be varied and quantified in bits relative to model capacity \citep{morris2026how}, whereas procedural tasks lack a comparable notion of load or capacity. 
Work that expresses explicit algorithms as transformer computations suggests possible resource measures such as depth and head count \citep{pmlr-v139-weiss21a,lindner2023tracr}, although these do not yet quantify how much capacity a learned procedure occupies in a model. Developing such measures would allow procedural load to be varied systematically as the declarative load, and test how task complexity and composition shape plasticity and stability.

Overall, pursuing these finer structural and functional resolutions may ultimately enable adaptation strategies that allocate updates according to \emph{where} and \emph{what} a model needs to learn, supporting more efficient, continually evolving AI systems.

\section{Conclusion}

We show that different transformer modules specialize differently in acquiring and preserving declarative and procedural capabilities, and that plasticity does not mirror stability. Attention projections are most plastic for procedural learning, whereas MLPs are most plastic for declarative memorization, yet modules with similar plasticity can differ sharply in retention. Reset-and-recovery experiments indicate that plasticity follows a module’s suitability for a task’s primary computation or storage, while stability depends on the broader roles that existing capabilities rely on, including access to information stored elsewhere. These asymmetric dynamics largely carry over to realistic tasks, motivating continual learning methods that consider both task type and module function.


\section{AI Use Statement}
We used AI to assist with polishing the paper writing, identifying suitable and accessible datasets for our experiments, and implementing code based on our research designs and ideas to conduct experiments.
We have reviewed all AI-assisted work. We take full responsibility for all content and results of this work, including those produced with AI assistance.


\bibliography{iclr2027_conference}
\bibliographystyle{iclr2027_conference}

\clearpage
\appendix

\section{Related Work}

\paragraph{Plasticity and stability in continual learning.}
The trade-off was first documented as catastrophic interference \citep{MCCLOSKEY1989109} and is typically addressed at the model level with task-agnostic remedies such as regularization \citep{Kirkpatrick_2017} and replay \citep{rolnick2019experiencereplaycontinuallearning,chaudhry2019tinyepisodicmemoriescontinual}.
For language models, common remedies include parameter-efficient adaptation \citep{hu2022lora} and self-distillation \citep{yang2024selfdistillation,shenfeld2026selfdistillation}, whose effects vary across tasks: LoRA forgets little on code and math \citep{biderman2024lora} yet degrades QA when injecting facts \citep{pletenev2025knowledge}.
Moreover, much measured forgetting may reflect lost access rather than lost knowledge: \citet{kotha2024understandingcatastrophicforgettinglanguage} attribute it to a shift in the task inferred by the model, while \citet{zheng2025spurious} show that lost task alignment can be restored with a small amount of old-task data.
Loss of plasticity over long task sequences \citep{Dohare2024LossOP} is orthogonal to the single-step setting we study.
These works largely treat the model as a whole; in contrast, we localize the access/storage distinction to modules: updating an access module (VO) yields forgetting that light replay largely reverses, whereas updating a storage module (MLP) yields forgetting that it only partly reverses (Section~\ref{sec:connect}).

\paragraph{Module-restricted adaptation.}
Studies that restrict fine-tuning to a subset of modules have typically emphasized one side of the plasticity--stability trade-off.
Attention-only fine-tuning is sufficient to adapt vision transformers \citep{touvron2022thingsknowvisiontransformers} and can teach vision-language models new skills with little forgetting \citep{zhu2025teach}, whereas feed-forward layers can be more effective targets for parameter-efficient learning \citep{he2022unifiedviewparameterefficienttransfer,schulman2025lora}; LoRA and full fine-tuning can also converge to different solutions \citep{shuttleworth2025lora}.
Taken separately, these results can appear conflicting because they concern different properties and task types; we instead measure both plasticity and stability across two task families and explain the resulting pattern through the distinct roles played by each module.

\paragraph{Functional specialization of transformer modules.}
Static interpretability work localizes factual associations to feed-forward key-value memories \citep{geva2021transformer,meng2022locating}, links induction-based in-context learning to attention circuits \citep{elhage2021mathematical,olsson2022context}, and shows that attention heads extract stored facts \citep{geva-etal-2023-dissecting}.
However, causal localization does not necessarily identify the best target for intervention: where causal tracing locates a fact does not predict which MLP layer is best to edit \citep{hase2023doeslocalizationinformediting}.
Controlled synthetic testbeds isolate architectural capabilities under clean measurement \citep{Allenzhu2025-canon}, while information-theoretic work quantifies memorization in bits per parameter \citep{morris2026how}.
We build on these tools to ask a dynamic question: how specialization shapes learning and retention under module-wise updates.
We probe module roles through reset-and-recovery, which perturbs module parameters and measures capability recovery after retraining, and use the declarative/procedural distinction from memory research \citep{cohen1980procedural,squire1992memory,mcclelland1995why,knowlton1996dissociation} as the organizing lens.


\section{Reproducibility Statement} 
We ensure reproducibility by providing an anonymous repository (\href{https://anonymous.4open.science/r/ICLR27_Plasticity-Stability-98B7/README.md}{link}) with source code, datasets, evaluation scripts used in this work.


\section{Additional Details of Synthetic Experiments}
\label{app:dfs_vs_mem_load50}

\paragraph{Training Setup}
Unless otherwise specified, we use an effective batch size of 256 and optimize with AdamW ($\beta_1=0.9$, $\beta_2=0.95$) with a weight decay of $0.01$. The learning rate is linearly warmed up over the first $2\%$ of training to a peak value of $3\times10^{-4}$, followed by cosine decay to $10\%$ of the peak learning rate. We apply gradient clipping with a maximum norm of $1.0$ and use \texttt{bfloat16} mixed-precision training.

\begin{figure}[t]
    \centering
    \begin{minipage}[t]{0.49\linewidth}
        \centering
        \includegraphics[width=\linewidth]{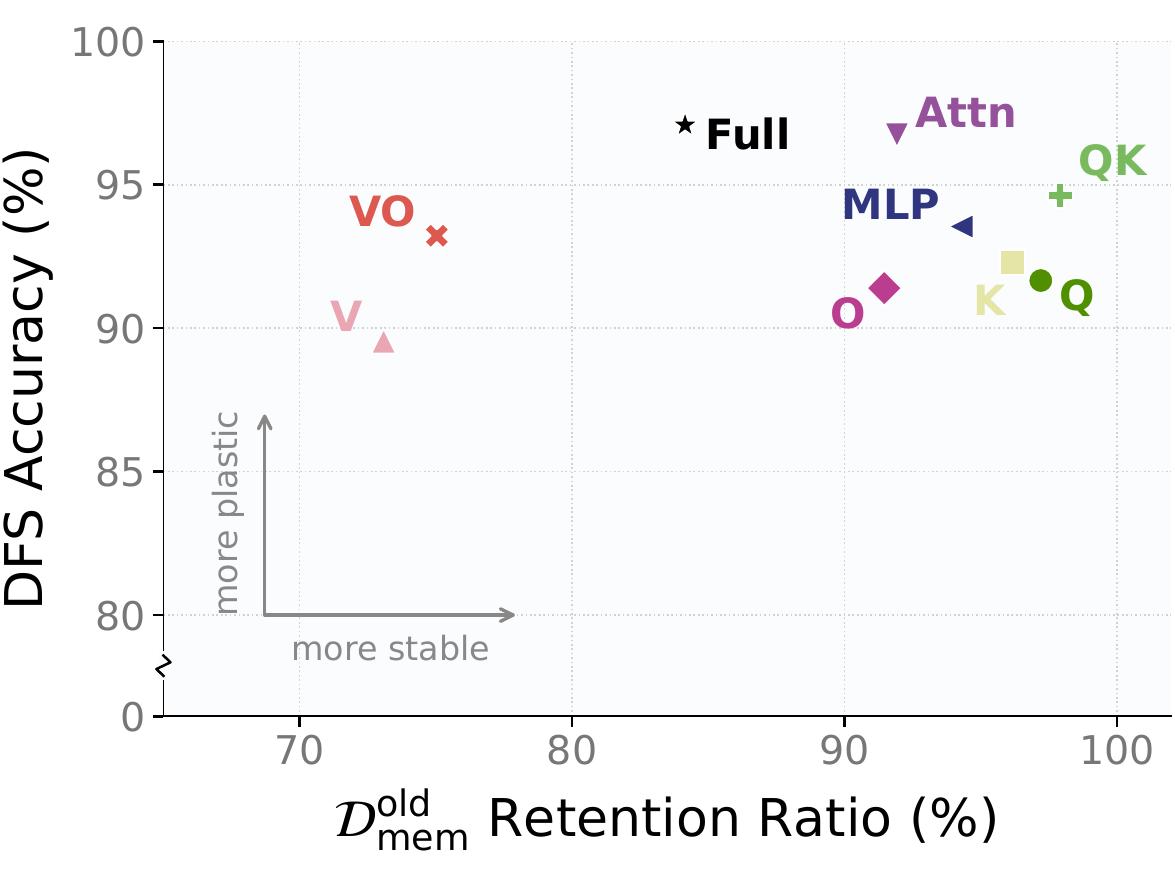}
        \vspace{-0.8em}
        
        {\small (a) Procedural adaptation}
    \end{minipage}
    \hfill
    \begin{minipage}[t]{0.49\linewidth}
        \centering
        \includegraphics[width=\linewidth]{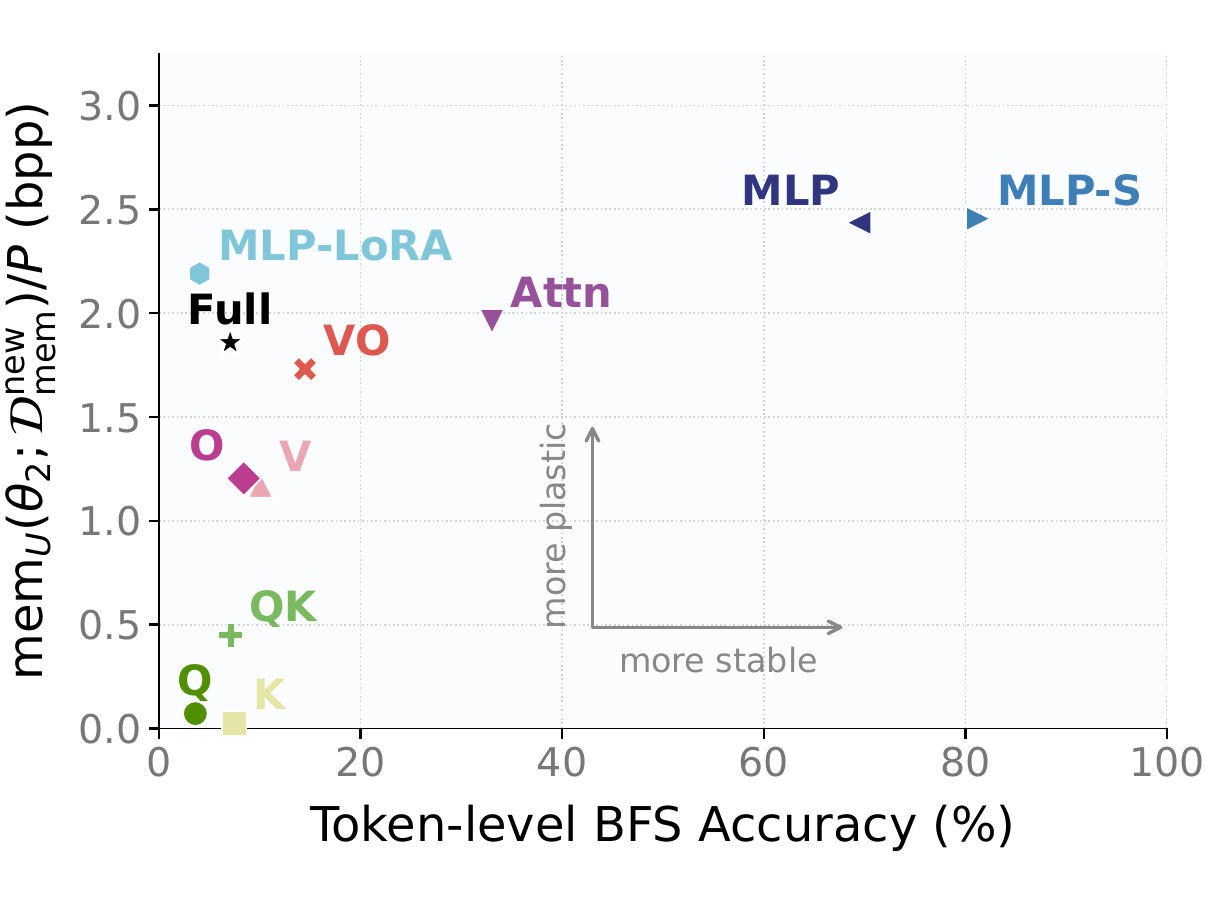}
        \vspace{-0.8em}
        
        {\small (b) Declarative adaptation}
    \end{minipage}
    \caption{
    \textbf{Plasticity--stability trade-offs under a non-saturated Stage-1 memory load.}
    We reduce the Stage-1 declarative memory load to 50\% of the estimated full-model capacity, at which the MLP alone is theoretically sufficient to store the entire memory bank.
    \textbf{(a)} DFS plasticity versus retention of the Stage-1 declarative memory after procedural adaptation.
    \textbf{(b)} New declarative memory acquired versus retained token-level BFS accuracy after declarative adaptation.
    The qualitative module-wise patterns closely match those observed under the saturated-load setting in the main text.
    }
    \label{fig:load50}
\end{figure}

\paragraph{Results under a Non-Saturated Stage-1 Memory Load}
Our main synthetic experiments deliberately use a heavily loaded Stage-1 checkpoint, where the declarative memory bank is chosen to saturate the model's memorization capacity.
To verify that the observed module-wise plasticity--stability patterns are not specific results of this saturation setup, we repeat both Stage-2 adaptation directions with a substantially lighter Stage-1 memory load, using only 50\% of the estimated full-model capacity.
At this load, the memory bank is small enough that the MLP alone is theoretically capable of storing all declarative information, removing the need to distribute storage across modules purely because of capacity constraints.
As shown in Figure~\ref{fig:load50}, the resulting trends closely match those in the main experiments.
Under procedural adaptation, QK remains substantially more stable than VO while achieving comparable DFS performance, and MLP-based updates still preserve more memory than VO.
In the reverse direction, when learning new declarative memory, MLP and MLP-S again provide the strongest combination of memory acquisition and BFS retention, whereas attention projections, particularly QK, acquire much less new memory despite preserving less of the previous procedural skill.
Thus, the qualitative plasticity--stability asymmetries reported in the main text persist even when Stage~1 does not saturate model memory, suggesting that they reflect module-dependent functional roles regardless of the capacity saturation.

\section{Additional Details of Realistic Experiments}
\label{app:real-data}

\begin{figure}[h]
    \centering
    \includegraphics[width=0.95\linewidth]{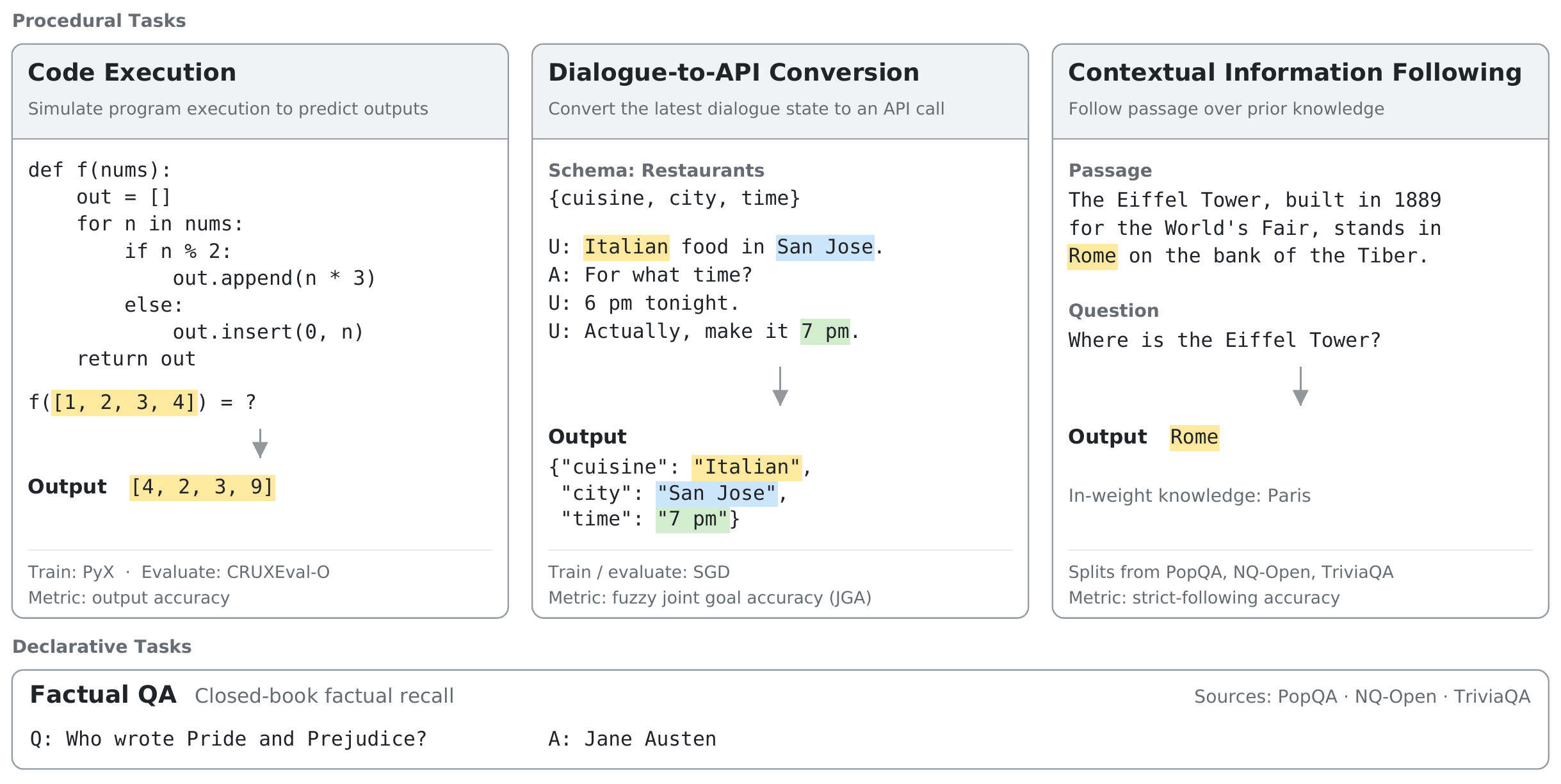}
    \caption{Realistic Procedural and Declarative Tasks}
    \label{fig:real_tasks}
\end{figure}

Table~\ref{tab:real-datasets} lists the number of samples used for training and evaluation in the realistic setting.
The code-execution training rows are PyX programs whose target is the program output, without a reasoning trace.
SGD training uses 20{,}000 user-turn service frames drawn from the official training split, and the SGD test set contains 11{,}539 frames of services seen during training and 34{,}577 frames of unseen services.
The context-following training set contains 140 conflict passages whose answer has been substituted, 700 reading passages for questions the base model cannot answer closed-book, and 96 passages that agree with the model's knowledge, built from TriviaQA and SQuAD; its evaluation pairs each of the 1{,}683 PopQA facts the base model knows with a one-sentence passage that states a different answer of the same relation.
The injected facts are 5{,}000 TriviaQA and 5{,}000 NQ-Open training questions and 4{,}966 PopQA questions; none overlaps an evaluation set, and the 33 facts the base model already answered correctly were removed.
After injection, the SGD skill is measured on the 17{,}283 test frames the model solved before injection.

\begin{table}[htbp]
\centering\small
\caption{Number of samples used for training and evaluation in the realistic setting. For the retention benchmarks, the count is the number of items the untouched base model answers correctly, out of the full benchmark.}
\label{tab:real-datasets}
\begin{tabular}{lllr}
\toprule
Task & Split & Data & Samples \\
\midrule
Code execution & train & PyX & 22{,}666 \\
 & eval & CRUXEval-O & 800 \\
\addlinespace
Dialogue-to-API conversion & train & SGD user-turn service frames & 20{,}000 \\
 & eval & SGD test frames & 46{,}116 \\
\addlinespace
Context following & train & TriviaQA and SQuAD passages & 936 \\
 & eval & PopQA conflict passages & 1{,}683 \\
\addlinespace
Knowledge injection & train & TriviaQA, NQ-Open, and PopQA facts & 14{,}966 \\
 & eval & injected facts, closed-book & 14{,}966 \\
 & eval & SGD test frames & 46{,}116 \\
\midrule
Knowledge retention & eval & PopQA & 1{,}683 / 14{,}267 \\
 & eval & TriviaQA (validation) & 535 / 11{,}313 \\
 & eval & NQ-Open (development) & 157 / 3{,}610 \\
\bottomrule
\end{tabular}
\end{table}

\noindent Below is the source of each dataset; the exact revisions and file hashes are recorded in the repository.
\begingroup\raggedright\def\UrlBreaks{\do\/\do-\do\_}
\begin{itemize}
\item PyX: \url{https://huggingface.co/datasets/semcoder/PyX}
\item CRUXEval: \url{https://github.com/facebookresearch/cruxeval}
\item SGD: \url{https://github.com/google-research-datasets/dstc8-schema-guided-dialogue}
\item SQuAD v1.1: \url{https://huggingface.co/datasets/rajpurkar/squad}
\item TriviaQA: \url{https://huggingface.co/datasets/mandarjoshi/trivia_qa}
\item NQ-Open: \url{https://github.com/google-research-datasets/natural-questions/tree/master/nq_open}
\item PopQA: \url{https://github.com/AlexTMallen/adaptive-retrieval}
\end{itemize}
\endgroup


\paragraph{Training Setup} All realistic experiments fine-tune the base model Qwen2.5-0.5B (\url{https://huggingface.co/Qwen/Qwen2.5-0.5B}) with MS-Swift.
Only the weights of the selected module are updated and every other parameter stays frozen: QK updates the query and key projections, VO the value and output projections, Attn all four attention projections, MLP the gate, up, and down projections, and Full all attention and MLP projections. Embeddings, normalization layers, and the tied output head stay frozen in every setup.
All runs use AdamW with a peak learning rate of $10^{-4}$, 3\% linear warm-up followed by cosine decay, weight decay 0.01, gradient clipping at 1.0, an effective batch size of 16, FP32 master weights with BF16 computation, and a loss on the answer tokens only.

\paragraph{Task Metrics}
CRUXEval-O counts a prediction as correct when the predicted output equals the reference under Python-literal equality.
SGD uses the official fuzzy joint goal accuracy, averaged over service frames.
For Knowledge Conflict, the strict follow rate is the fraction of PopQA conflict passages for which the first output line equals the answer stated in the passage.
For knowledge injection, facts learned is the strict closed-book exact match on the injected facts, and SGD skill retained is the fraction of the 17{,}283 test frames solved before injection that the model still solves completely.

\paragraph{Knowledge Retention}
Retention is measured on the benchmark items that the untouched base model answers correctly under both of two fixed question orders, the source order and the reversed order rotated by one. An item counts as retained if the fine-tuned model still answers it correctly under both orders.
The prompt is \texttt{Question:\ \{question\}\textbackslash{}nAnswer:} without a chat template or demonstrations, decoding is greedy in FP32, and PopQA, TriviaQA, and NQ-Open answers are scored on the first output line with each benchmark's original normalization and alias matching.
For context following, the PopQA, TriviaQA, and NQ-Open retention scores use a single pass with lenient containment matching, on the items the base model answers correctly under the same prompt (1445, 535, and 156 items, respectively).

\end{document}